\documentclass[10pt]{article} 
\usepackage[utf8]{inputenc} 

\usepackage{geometry} 
\usepackage{graphicx} 
\usepackage{url}

\usepackage{amssymb,amsfonts,amsmath,amsthm}
\usepackage{booktabs} 
\usepackage{array} 
\usepackage{paralist} 
\usepackage{verbatim} 
\usepackage{subfig} 
\usepackage{color}
\usepackage{cite}
\usepackage{epsfig}
\usepackage[labelfont=bf]{caption}

\usepackage{fancyhdr} 
\usepackage{sectsty}
\allsectionsfont{\sffamily\mdseries\upshape} 

\usepackage[nottoc,notlof,notlot]{tocbibind} 
\usepackage[titles,subfigure]{tocloft} 

\usepackage{amssymb,amsfonts,amsmath}
\newtheorem{theorem}{Throrem}
\newtheorem{lemma}{Lemma}
\newcommand{\argmax}{\operatornamewithlimits{argmax}}

\begin{document}
\title{\bf{Ranking and combining multiple predictors without labeled data}}
\author{Fabio Parisi $^{1,\#}$, Francesco Strino $^{1,\#}$, Boaz Nadler$^{2}$, Yuval Kluger $^{1, 3, \ast}$}
\date{}
\maketitle
\footnotesize{\noindent{$^{1}$ Yale University School of Medicine, Department of Pathology, 333 Cedar St., New Haven, CT 06520, USA. $^{2}$ Weizmann Institute of Science, Department of Computer Science and Applied Mathematics, Rehovot, 76100 Israel. $^{3}$ NYU Center for Health Informatics and Bioinformatics New York University Langone Medical Center 227 East 3030$^{\text{th}}$ Street, New York, NY 10016, USA.
$^{\#}$ These authors contributed equally to this work.$^{\ast}$ E-mail: yuval.kluger@yale.edu.}

\begingroup
\renewcommand{\addcontentsline}[3]{}

\section*{Abstract} 
In a broad range of classification and decision making problems, one is given the advice or predictions of several classifiers, of unknown reliability, over multiple questions or queries. This scenario is different from the standard supervised setting, where each classifier accuracy can be assessed using available labeled data, and raises two questions: given only the predictions of several 
classifiers over a large set of unlabeled test data, is it possible to a) reliably rank them; and b) construct a meta-classifier more accurate than most classifiers in the ensemble?

Here we present a novel spectral approach to address these questions. First, assuming conditional independence between classifiers, we show that the off-diagonal entries of their covariance matrix correspond to a rank-one matrix. Moreover, the classifiers can be ranked using the leading eigenvector of this covariance matrix, as its entries are proportional to their balanced accuracies. Second, via a linear approximation to the maximum likelihood estimator, we derive the Spectral Meta-Learner (SML), a novel ensemble classifier whose weights are equal to this eigenvector entries. On both simulated and real data, SML typically achieves a higher accuracy than most classifiers in the ensemble and can provide a better starting point than majority voting, for estimating the maximum likelihood solution. Furthermore, SML is robust to the presence of small malicious groups of classifiers designed to veer the ensemble prediction away from the (unknown) ground truth.

\section*{Introduction}
Everyday, multiple decisions are made based on input and suggestions from several sources, either algorithms or advisers, of unknown reliability. 
Investment companies handle their portfolios by combining reports from several analysts, each providing recommendations on buying, selling or holding multiple stocks \cite{tsai_combining_2010, zweig_making_2011}. 
Central banks combine surveys of several professional forecasters to monitor rates of inflation, real GDP growth and unemployment \cite{european_central_bank_survey, federal_reserve_bank_of_philadelphia_survey,monetary_authority_of_singapore_survey, genre_combining_2010}. 
Biologists study the genomic binding locations of proteins, by combining or ranking the predictions of several peak detection algorithms applied to large-scale genomics data \cite{micsinai_picking_2012}. 
Physician tumor boards convene a number of experts from different disciplines to discuss patients whose diseases pose diagnostic and therapeutic challenges \cite{wright_multidisciplinary_2007}. 
Peer review panels discuss multiple grant applications and make recommendations to fund or reject them \cite{cole_peer_1978}. 
The examples above describe scenarios in which several human advisers or algorithms, provide their predictions or answers to a list of queries or questions. A key challenge is to improve decision-making by combining these multiple predictions, of unknown reliability. Automating this process, of combining multiple predictors, is an active field of research in decision science (\url{cci.mit.edu/research}), medicine~\cite{margolin_systematic_2013}, business (\cite{linstone_delphi},~\cite{stock_forecasting_2006} and \url{www.kaggle.com/competitions}) and government (\url{www.iarpa.gov/Programs/ia/ACE/ace.html} and \url{www.goodjudgmentproject.com}), as well as in statistics and machine learning.

Such scenarios, whereby advisers of unknown reliability provide potentially conflicting opinions, or propose to take opposite actions, raise several interesting questions: How should the decision-maker proceed to identify who, among 
the advisers, is the most reliable? Moreover, is it possible for the decision-maker to cleverly combine the collection of answers from all the advisers and provide even more accurate answers?

In statistical terms, the first question corresponds to the problem of estimating prediction performances of pre-constructed classifiers 
({e.g.}, the advisers) in absence of class labels. Namely, each classifier was constructed independently on a potentially different training dataset ({e.g.}, each adviser trained on his/her own using possibly different sources of information), yet they are all being applied to the same new test data ({e.g.}, list of queries) for which labels are not available, either because they are expensive to obtain, or because they will only be available in the future, after the decision has been made. 
In addition, the accuracy of each classifier on its own training data is unknown. This scenario is markedly different from
the standard supervised setting in machine learning and statistics. There, classifiers are typically trained on 
the same labeled data, and can be ranked, for example, by comparing their empirical accuracy on a common labeled validation set. In this paper we show that under standard assumptions of independence between classifier errors, 
their unknown performances can still be ranked even in the absence of labeled data.

The second question raised above corresponds to the problem of combining predictions of pre-constructed classifiers to form a meta-classifier with improved prediction performance. This problem arises in many fields, including  combination of forecasts in decision science and crowdsourcing in machine learning, which have each derived different approaches to address it. If we had external knowledge or historical data to assess the reliability of the available classifiers we could use well-established solutions relying on panels of experts or forecast combinations 
\cite{linstone_delphi, stock_forecasting_2006,timmermann_chapter_2006,DeGroot}. 
In our problem such knowledge is not always available and thus these solutions are in general not applicable. The oldest solution that does not require additional information is {\it majority voting}, whereby the predicted class label is determined by a rule of majority, with all advisers assigned the same weight.  

More recently, iterative likelihood maximization procedures, pioneered by Dawid and Skene \cite{dawid_maximum_1979}, have been proposed, in particular in crowdsourcing applications  \cite{karger_budget-optimal_2011, karger_iterative_2011, whitehill_whose,padhraic_smyth_inferring_1996, sheng_get_2008, welinder_multidimensional_2010, yan_modeling_2010, raykar_learning_2010}. Due to the non-convexity of the likelihood function, these techniques often converge only to a local, rather than global, maximum and require careful initialization. Furthermore, there are typically no guarantees on the quality of the resulting solution.

In this paper we address these questions via a novel spectral analysis, 
that yields four major insights:

\begin{enumerate}
\item Under standard assumptions of independence between classifier errors, in the limit of an infinite test set, the off-diagonal entries of the population covariance matrix of the classifiers correspond to a {\it rank-one matrix}.

\item The entries of the leading eigenvector of this rank-one matrix are proportional to the {\it balanced accuracies} of the classifiers. Thus, a spectral decomposition of this rank-one matrix provides  a  computationally efficient approach to rank the performances of an ensemble of classifiers. 

\item A linear approximation of the maximum likelihood estimator yields an ensemble learner whose weights are proportional to the entries of this eigenvector. This represents a new, easy to construct, unsupervised ensemble learner, which we term Spectral Meta-Learner (SML).

\item An interest group of 
conspiring classifiers (a {\it cartel}), which maliciously attempts to veer the overall ensemble solution away from the (unknown) ground truth, leads to a {\em rank-two} covariance matrix. Furthermore, in contrast to majority voting, SML is robust to the presence of a small enough cartel whose members are unknown.

\end{enumerate}

In addition,  we demonstrate the advantages of spectral approaches based on these insights, using both simulated and real-world datasets.
When the independence assumptions hold approximately, SML is typically better than most classifiers in the ensemble and their majority vote, achieving results comparable to the maximum likelihood estimator (MLE). Empirically, we find SML to be a better starting point for computing the MLE that consistently leads to improved performance. Finally, spectral approaches are also robust to cartels and therefore helpful in analyzing surveys where a biased sub-group of advisers (a cartel) may have corrupted the data. 

\section*{Problem setup}
For simplicity, we consider the case of questions with yes/no answers. 
Hence, the advisers, or algorithms, provide to each query only one of two possible answers, either $+1$ (positive) or $-1$ (negative). 
Following standard statistical terminology, the advisers or algorithms  are called {\it binary classifiers}, and their answers are termed predicted {\it class labels}.
Each question is represented by a feature vector $x$ contained in a feature space $\mathcal X$. 

In detail, let $\{f_{i}\}_{i=1}^M$ be $M$ binary classifiers of unknown reliability, each providing predicted class labels \(f_{i}(x_k)\)
to a set of $S$ instances $D=\{x_k\}_{k=1}^S\subset \mathcal X$, whose vector of true (unknown)
class labels is denoted by ${\bf y}=(y_1,\ldots,y_S)$.
We assume that each classifier $f_i:\mathcal X\to\{-1,1\}$  was trained in a manner undisclosed to us using its own labeled training set, which is also unavailable to us. Thus, we view each classifier as a black-box function of unknown classification accuracy. 

Using only the predictions of the $M$ binary classifiers on the unlabeled set $D$ and without access to any labeled data,
we consider the two problems stated in the introduction: 
i) rank the performances of the $M$ classifiers;
and ii) combine their predictions to provide an improved estimate  $\hat{\bf y}=(\hat{y}_1,\ldots,\hat{y}_S)$ of the true class label vector ${\bf y}$.

We represent an instance and class label pair $(X,Y)\in\mathcal X \times\{-1,1\}$ as a random vector 
with probability density function $p(x,y)$, and with marginals \(p_{X}(x)\) and \(p_{Y}(y).\) 

In the present study, we measure the performance of a 
binary classifier \(f\) by its  \textit{balanced accuracy}
$\pi$, defined as  
\begin{align}
\pi&=
\frac{\mbox{sensitivity + specificity}}{2} = \frac12\left(\psi+\eta\right),
\end{align}
where $\psi$ and $\eta$ are its \textit{sensitivity} (fraction of correctly predicted positives) and 
\textit{specificity} (fraction of correctly predicted negatives). 
Formally, these quantities are defined as  

\begin{equation}\label{MT-PSIETA}
\psi = \Pr[f(\!X\!)\!=\!Y|Y\!=\!1],\mbox{ and}\quad \eta =\Pr[f(\!X\!)\!=\!Y|Y\!=\!-\!1].
\end{equation}

\subsection*{Assumptions}

In our analysis we make the following two assumptions: i) The $S$ unlabeled instances $x_k\in D$ are i.i.d. realizations from the marginal distribution $p_{X}(x)$; and ii) the $M$ classifiers are conditionally independent, in the sense that prediction errors made by one classifier are independent of those made by any other classifier. Namely, for all $1\leq i\neq j\leq M$, and for each of the two class labels, with $ a_i, a_j \in\{-1,1\}$ 
\begin{equation}\label{MT-StatIndependence}
\Pr[f_i(\!X\!)\!=\!a_i,f_j(\!X\!)\!=\!a_j|Y]=\Pr[f_i(\!X\!)\!=\!a_i|Y]\Pr[f_j(\!X\!)\!=\!a_j|Y].
\end{equation}
Classifiers that are nearly conditionally independent may arise, for example, from advisers who did not communicate with each other, or from algorithms that are based on different design principles or independent sources of information.
Note that these assumptions appear also in other works considering a setting similar to ours \cite{dawid_maximum_1979,raykar_learning_2010}, as well as in supervised learning, in the development of classifiers ({e.g.}, Na\"ive Bayes) and ensemble methods \cite{dietterich_ensemble_2000}.

\section*{Ranking of classifiers}
To rank the $M$ classifiers without any labeled data, in this paper we present a spectral approach based on the covariance matrix of the $M$ classifiers. To motivate our approach it is instructive to first study its asymptotic structure as the number of unlabeled test data tends to infinity, $|D|=S\to\infty$. Let $Q$ be the $M\times M$ population covariance matrix of the $M$ classifiers, whose entries are defined as

\begin{equation}
q_{ij} = \mathbb{E}[(f_i(X)-\mu_i)(f_j(X)-\mu_j)]
\end{equation}
where $\mathbb{E}$ denotes expectation with respect to the density 
$p(x,y)$ and $\mu_i = \mathbb{E}[f_i(X)]$. 

The following lemma, proven in the supplementary information, characterizes the relation between the matrix $Q$ and the balanced accuracies of the $M$ classifiers:  

\begin{lemma}\label{MT-LEMMA1} The entries $q_{ij}$ of $Q$ are equal to 
\begin{equation}\label{MT-eqCov}
q_{ij} = \left\{
  \begin{array}{c c l}
    1-\mu_i^2 & &  i=j\\
    (2\pi_i-1)(2\pi_j-1)\left(1-b^2\right) & & \text{\rm otherwise}\\
  \end{array} \right. 
\end{equation}
where $b\in(-1,1)$ is the class imbalance, 
\begin{equation}
b = \Pr[Y=1]-\Pr[Y=-1]. 
        \label{MT-eq:class_imbalance}
\end{equation}
\end{lemma}

\hfill

The key insight from this lemma is that the off-diagonal entries 
of $Q$ are identical to those of a \textit{rank-one matrix} 
\(R = \lambda {\bf v}{\bf v}^T\) with unit-norm eigenvector \({\bf v}\)
and eigenvalue 
\begin{equation}
\lambda=(1-b^2)\cdot\sum_{i=1}^M (2\pi_i-1)^2.
        \label{MT-eq:lambda_tilde_Q}
\end{equation}
Importantly, up to a sign ambiguity, the entries of  ${\bf v}$
 are \textit{proportional} to the balanced accuracies of the $M$ classifiers,
\begin{equation}
v_i\propto (2\pi_i-1).
\end{equation}
Hence, the $M$ classifiers can be ranked according to their balanced accuracies by sorting the entries of the eigenvector \({\bf v}\). 

While typically neither $Q$ nor ${\bf v}$ are known, both can be
estimated from the finite unlabeled dataset \(D\). We denote the corresponding sample covariance matrix by $\hat Q$. Its entries are 
\[
\hat q_{ij}=\frac1{S-1}\sum_{k=1}^S (f_i(x_k)-\hat\mu_{i})(f_j(x_k)-\hat\mu_j)
\]
where \(\hat\mu_i=\frac1S\sum_k f_i(x_k).\) 
Under our assumptions, $\hat Q$ is an unbiased estimate of $Q$, $\mathbb{E}[\hat Q] = Q$.
Moreover, the variances of its off-diagonal entries are given by
\begin{equation}\label{MT-eqvarQ}
Var[\hat{q}_{ij}]= \frac{(1-{\mu}_i^2)\cdot (1-{\mu}_j^2)}{S-1} + 
\frac{{q}_{ij}}{S}\left(4{\mu}_i{\mu}_j - \frac{S-2}{S-1}{q}_{ij} \right).
\end{equation}
In particular, $\hat q_{ij} - q_{ij}=O(1/\sqrt{S})$ and asymptotically $\hat Q\to Q$ as $S\to\infty$.
Hence, for a sufficiently large unlabeled set \(D\), it should be possible to accurately estimate from $\hat Q$ the eigenvector $\mathbf{v}$ and consequently the ranking of the \(M\) classifiers.

One possible approach is to construct an estimate \(\hat{R}\) of the rank one matrix \(R\) and then compute its leading eigenvector. Given that \(\mathbb{E}[\hat Q] = Q\), for all \(i\neq j\) we may estimate \(\hat r_{ij}=\hat q_{ij}\), and we only need to estimate the diagonal entries of \(R\).  
A computationally efficient way to do this, by solving a set of linear equations, is based on the following observation: upon the change of variables $ r_{ij} = e^{t_i} \cdot e^{t_j}$, we have for all \(i\neq j\), 
\[
\log |r_{ij}|  - t_i - t_j =
\log |q_{ij}|  - t_i - t_j = 0.
\]
Hence, if we knew $q_{ij}$ we could find the vector ${\bf t}$ by solving the above system of equations.
In practice, as we only have access to $\hat q_{ij}$ we thus look for a vector $\hat{\bf t}$ with small residual 
error in the above $M(M-1)/2$ equations.
We then estimate the diagonal entries by ${\hat{r}}_{ii}=\exp(2\hat t_{i})$ and proceed with eigendecomposition of $\hat R$.  
Further details on this and other approaches to estimate $\mathbf{v}$
appear in the Supplementary Information.

Next, let us briefly discuss the error in this approach. First, since  $\hat Q\to Q$ as \(S\to\infty\), it follows that $\hat {\bf t}\to {\bf t}$
and consequently $\hat R\to R$. Hence, asymptotically we perfectly recover the correct ranking of the $M$ classifiers. 
Since $R$ is rank-one, $\hat R-R = O(1/\sqrt{S})$ and both \(R\)
and \(\hat R\)
are symmetric, as shown in the Supplementary Information, 
the leading eigenvector is stable to small perturbations. In particular, $\hat {\bf v}-{\bf v}= O(\frac1\lambda\frac1{\sqrt{S}}).$ Finally, note that if all classifiers are better than random and the class imbalance is bounded away from $\pm1$, then we have a large spectral gap with $\lambda = O(M)$. 

\section*{The Spectral Meta Learner (SML)}

Next, we turn to the problem of constructing a meta-learner expected to be more accurate than most (if not all) of the \(M\) classifiers in the ensemble. In our setting, this is equivalent to estimating the \(S\) unknown labels \(y_1,\ldots,y_S\) by combining the labels predicted by the $M$ classifiers.

The standard approach to this task is to determine for all the unlabeled instances the maximum likelihood estimator (MLE) $\mathbf{\hat{y}^{\text{ML}}}$ of their true class labels $\mathbf{y}$ \cite{dawid_maximum_1979}. 
Under the assumption of independence between classifier errors and between instances, the overall likelihood is the product of the likelihoods of the \(S\) individual  instances, where the likelihood of a label \(y\) for an instance \(x\) is 

\begin{equation}
\mathfrak{L}(f_1(x),\ldots, f_M(x);y) = \\ \prod_{i=1}^{M}\Pr(f_i(x)|y).
\end{equation}

As shown in the Supplementary Information, the MLE can be written as a weighted sum of the binary labels $f_i(x) \in \{-1,1\}$, with weights that depend on the sensitivities $\psi_i$ and specificities $\eta_i$ of the classifiers. For an instance \(x\), 
\begin{eqnarray}\label{MT-MLE}
\hat{y}^{\text{(ML)}} & = & \argmax_{y}   \mathfrak{L}(f_{1}(x),\ldots,f_M(x);y) 
       \nonumber \\ 
  & = &  \text{sign}\Big(\sum_{i=1}^M f_i(x) \log \alpha_i
       + \log \beta_i \Big)
        \label{MT-eq:y_ML}
 \end{eqnarray}
where
\begin{equation}
\alpha_i = \frac{\psi_{i}\eta_{i}}{(1-\psi_{i})(1-\eta_{i})}, 
\ \ \ \beta_i  = \frac{\psi_{i}(1-\psi_{i})}{\eta_{i}(1-\eta_{i})}.
        \label{MT-eq:alpha_beta}
\end{equation}
Eq.~\eqref{MT-eq:y_ML} shows that the MLE is a \textit{linear ensemble classifier}, whose weights depend, unfortunately, on the unknown specificities and sensitivities of the \(M\) classifiers. 

The common approach, pioneered by Dawid and Skene \cite{dawid_maximum_1979}, is to look for all \(S\) labels and $M$ classifier specificities and sensitivities that {\em jointly} maximize the likelihood. Given an estimate of the true class labels, it is straightforward to estimate each classifier sensitivity and specificity. 
Similarly, given estimates of \(\psi_i$ and $\eta_i \), the corresponding estimates of $\mathbf{y}$ are easily found via \eqref{MT-eq:y_ML}. Hence, 
the MLE is typically approximated by expectation-maximization (EM) \cite{whitehill_whose, padhraic_smyth_inferring_1996, sheng_get_2008, welinder_multidimensional_2010, raykar_learning_2010}. 

As is well known, the EM procedure is guaranteed to increase the likelihood at each iteration till convergence. However, its key limitation is that due to the non-convexity of the likelihood  function, the EM iterations often converge to a local (rather than global) maximum. 

Importantly, the EM procedure requires  an initial guess of the true labels \(\mathbf{y}\). A common choice is the simple majority rule of all classifiers. As noted in previous studies, majority voting may be suboptimal, and starting from it, the EM procedure may converge to suboptimal local maxima \cite{raykar_learning_2010}.  
Thus, it is desirable, and  sometimes crucial, to initialize the EM algorithm with an estimate $\mathbf{\hat{y}}$ that is close to the true label $\mathbf{y}$. 

Using the eigenvector described in the previous section, we now present a novel construction of an initial guess that is typically more accurate than majority voting. To this end, note that a Taylor expansion of the unknown coefficients \(\alpha_i\)
and $\beta_i$ in \eqref{MT-eq:alpha_beta} around \((\psi_i,\eta_i)=(1/2,1/2) \)
gives, up to second order terms $O((\psi_i-1/2)^2,(\eta_i-1/2)^2,(\psi_i-1/2)\cdot (\eta_i-1/2))$,
\begin{equation}
\alpha_i \approx 1 +4(\psi_i+\eta_i-1)=1+4(2\pi_i-1),\quad \beta_i\approx 1.
        \label{MT-eq:alpha_beta_expansion}
\end{equation}
Hence, combining  \eqref{MT-eq:alpha_beta_expansion} with a first order Taylor expansion of  the argument inside the sign function in  \eqref{MT-MLE}, around  $(\psi_i,\eta_i)=(1/2,1/2)$
yields
\begin{equation}\label{MT-MLEo2}
\hat{y}^{\text{(ML)}}_k \approx \text{sign}
\Big(\sum_{i=1}^M f_i(x_k)(2 \pi_i-1) 
\Big).
\end{equation}

Recall that by Lemma~\ref{MT-LEMMA1}, up to a sign ambiguity the entries of the leading eigenvector of \(R\)  are proportional to the balanced accuracies of the classifiers, $v_i \propto (2\pi_i - 1)$. 
This sign ambiguity can be easily resolved if we assume, for example, that most classifiers are better than random. Replacing \(2\pi_i-1\) in \eqref{MT-MLEo2} by the eigenvector entries $\hat v_i$ of an estimate of \(R\) yields a novel spectral-based ensemble classifier, which we term the Spectral Meta-Learner (SML), 
\begin{equation}\label{MT-SML}
\hat{y}^{\text{(SML)}}_k = \text{sign}
\Big(\sum_{i=1}^M f_i(x_k)\cdot \hat v_i\Big).
\end{equation}

Intuitively, we expect SML to be more accurate than majority voting as it attempts to give more weight to more accurate classifiers. 
Lemma~\ref{SI-lemma:SML_Vs_Voting} in the Supplementary Information provides insights on the improved performance achieved by SML in the special case when all algorithms but one have the same sensitivity and specificity. 
Numerical results for more general cases are described in the simulation section, where we also show that empirically, on several real data problems, SML provides a better initial guess than majority voting for EM procedures that iteratively estimate the MLE.

\section*{Learning in the Presence of a Malicious Cartel}
Consider a scenario whereby  a small fraction $r$ of  the \(M\) classifiers belong to a conspiring {\em cartel} (e.g., representing a junta or an interest group), maliciously designed to veer the ensemble solution toward the cartel's target and away from the truth. 
The possibility of such a scenario raises the following question: how sensitive are SML and majority voting to the presence of a cartel? In other words, to what extent can these methods ignore, or at least substantially reduce, the effect of the cartel classifiers without knowing their identity? 

To this end, let us first introduce some notation. Let the \(M\) classifiers be composed of a subset \(P$ of $(1-r)M\) ``honest" classifiers
and a subset $C$ of \(rM\) malicious cartel classifiers. The honest classifiers satisfy the assumptions of the previous section: each classifier attempts to correctly predict the truth with a balanced accuracy \(\pi_i\), and different classifiers make independent errors. The cartel classifiers, in contrast, attempt to predict a {\em different} target labeling, ${\bf T}$.
We assume that conditional on both the cartel's target and the true label, the classifiers in the cartel make independent errors. Namely, for all $i,j \in C$, and for any labels $a_i,a_j,Y,T \in \{-1,1\}$
\begin{equation}\label{CartelIndependence}
\Pr\left[f_i\!(\!X\!)\!=\!a_i,f_j\!(\!X\!)\!=\!a_j|T\!,\!Y\right]\!=\Pr[f_i\!(\!X\!)\!=\!a_i|T]\Pr[f_j\!(\!X\!)\!=\!a_j|T].
\end{equation}
Similarly to the previous sections, we assume that the prediction errors of cartel and honest classifiers are also (conditionally)\ independent. 

The following lemma, proven in the supplementary information, expresses the entries of the population 
covariance matrix \(Q\) in terms of the following quantities: the balanced accuracies of the $M$ classifiers, 
the balanced accuracy $\pi_c$ of the cartel's target with respect to 
the truth, and the balanced accuracies $\xi_j$ of the $r\!\cdot\!M$ cartel members relative to their target.  

\hfill

\begin{lemma}\label{MT-LEMMA3} Given \((1-r)M\) honest classifiers 
and $r M$ classifiers of a cartel $C$, the entries $q_{ij}$ of $Q$ 
satisfy 
\begin{equation}\label{MT-eqCovcart}
q_{ij} = \left\{
  \begin{array}{c c l}
    1-\mu_i^2 & &  i=j\\
    (2\pi_i-1)(2\pi_j-1)\left(1-b^2\right) & & i\in P, j\in P\\\
    (2\pi_i-1)(2\pi_c-1)(2\xi_j-1)\left(1-b^2\right) & &i\in P, j\in C\\\
    (2\xi_i-1)(2\xi_j-1)\left(1-b^2\right) & & i\in C, j\in C
  \end{array} \right. 
\end{equation}
where $b\in(-1,1)$ is the class imbalance, as in \eqref{MT-eq:class_imbalance}.
\end{lemma}

Next, the following theorem shows that in the presence of a single cartel, the off-diagonal entries of $Q$ correspond to a \textit{rank-two} matrix.
We conjecture that in the presence of \(k\) independent cartels, the respective rank is \((k+1)\).

\hfill

\begin{theorem}\label{MT-THEO1} Given $(1-r)M$ honest classifiers and \(rM\)
classifiers belonging to a cartel, $0<r<1$, the off-diagonal entries 
of $Q$ correspond to a {\bf rank-two matrix} with eigenvalues
\begin{equation}
\begin{array}{r c l}
\lambda_1&=&\lambda_P\cos^2\alpha
 + \lambda_C\sin^2\beta \\
\lambda_2&=& \lambda_P\sin^2\alpha
+ \lambda_C\cos^2\beta
\end{array}
\end{equation}
and eigenvectors
\begin{equation}
e_{1i}  \!=\!  \left\{
 \begin{array}{c l}
 \! (2\pi_i-1) \cos\alpha & i\in P\\
 \! (2\xi_i-1) \sin\beta  & i\in C\\
 \end{array} \right.
\ 
e_{2i} \!=\! \left\{
 \begin{array}{c l}
 \! (2\pi_i-1)\sin\alpha & i\in P\\
 \! (2\xi_i-1)\cos\beta & i\in C\\
 \end{array} \right. 
\end{equation}
where
\begin{equation}
\lambda_P=(1\!-\!b^2)\sum_{j\in P}(2\pi_j\!-\!1)^2,\quad
\lambda_C=(1\!-\!b^2)\sum_{j\in C}(2\xi_j\!-\!1)^2
\end{equation}
and, with $k_1 = 2\pi_c - 1$, $k_2 = \lambda_C/\lambda_P$,
\[
\alpha\! =\tfrac{1}{2}\arctan\left(\tfrac{k_1k_2}{k_2(1-2k_1^2)-1}\right)\!,\,
\beta \! = \tfrac{1}{2}\arctan\left(\tfrac{2k_1\sqrt{1-k_1^2}}{1 - k_2 - 2k_1^2}\right).
\]
\end{theorem}

\hfill

As an illustrative example of Theorem~\ref{MT-THEO1}, consider the case where the cartel's target is unrelated to the truth, i.e. $\pi_c = 1/2$. 
In this case $\alpha=\beta=0$, so $\lambda_1=\lambda_P$, $\lambda_2=\lambda_C$ and 
\begin{equation}
e_{1i}  =  \left\{
 \begin{array}{c l}
 2\pi_i-1 & i\in P\\
 0 & i\in C\\
 \end{array} \right.
 \quad
e_{2i}  = \left\{
 \begin{array}{c l}
 0 & i\in P\\
 2\xi_i-1 & i\in C\\
 \end{array} \right. 
\end{equation}
Next, according to \eqref{MT-SML} SML weighs each classifier by the corresponding entry in the leading eigenvector.
Hence, if the cartel's target is orthogonal to the truth \((\pi_c=1/2\)) and $\lambda_P > \lambda_C$, SML asymptotically
\textit{ignores} the cartel (Fig.~\ref{SI-fig_FIG1}).  In contrast, regardless of $\pi_c$, majority voting is affected by the cartel,  proportionally to its fraction size $r$. Hence, SML is more robust than majority voting to the presence of such a cartel.

\section*{Application to simulated and real-world datasets}
The examples provided in this section showcase strengths and limitations of spectral approaches to the problem of ranking and combining multiple predictors without access to labeled data. First, using simulated data of an ensemble of independent classifiers and an ensemble of independent classifiers containing one cartel, we confirm the expected high performance of our ranking and SML algorithms. In the second part we consider the predictions of 33 machine learning algorithms as our ensemble of binary classifiers, and test our spectral approaches on 17 real-world datasets collected from a broad range of application domains.

\subsection*{Simulations}
We simulated an ensemble of \(M=100\) independent classifiers providing predictions for \(S=600\) instances, whose ground truth had class imbalance \(b=0\). To imitate a difficult setting, where some classifiers are worse than random, each generated classifier had different sensitivity and specificity chosen at random such that its balanced accuracy was 
uniformly distributed in the interval $[0.3,0.8]$. We note that classifiers that are worse than random may occur in real studies, when the training data is too small in size or not sufficiently representative of the test data. Finally, we considered the effect of a malicious cartel consisting of \(33\%\) of the classifiers, having their own target labeling. More details about the simulations are provided in the Supplementary Information. 

{\em Ranking of Classifiers:} We constructed the sample covariance matrix, corrected its diagonal as described in the Supplementary Information and computed its leading eigenvector \(\hat{\bf v}\). In both cases (independent classifiers and cartel), with probability of at least \(80\%\), the classifier with highest accuracy was also the one with the largest entry (in absolute value) in the eigenvector \(\hat{\bf v} \), and with probability $>99\%$ its inferred rank was among the top five classifiers (Fig.~\ref{SI-fig_FIG2}). Note that even if the test data of size $S = 600$ were fully labeled, identifying the best performing classifier would still be prone to errors, as the estimated balanced accuracy has itself an error of \(O(1/\sqrt{S}).\)

{\em Unsupervised Ensemble-Learning:} Next, for the same set of simulations we compared the balanced accuracy of majority voting and of SML. We also considered the predictions of these two meta-learners as starting points for iterative EM calculation of the MLE (iMLE). 
As shown in Fig.~\ref{MT-fig:SimPerf}, SML was significantly more accurate than majority voting. Furthermore, applying an EM procedure with SML as an initial guess provided relatively small improvements in the balanced accuracy. Majority voting, in contrast, was less robust. Moreover, in the presence of a cartel, computing the MLE with majority voting as its starting point exhibited a multi-modal behavior, sometimes converging to a local maxima with a relatively low balanced accuracy.

A more detailed study of the sensitivity of SML and majority voting and their respective iMLE solutions versus the size of a malicious cartel with $\pi_c = 0.5$ is shown in Fig.~\ref{SI-fig_FIG6}. As expected, the average balanced accuracy of all methods decreases as a function of the cartel's fraction $r$, and once the cartel's fraction is too large all approaches fail. 
In our simulations, both SML and iMLE initialized with SML were far more robust to the size of the cartel than either majority voting or iMLE initialized with majority voting. With a cartel size of 20\%, SML was still able to construct a nearly perfect predictor, whereas the balanced accuracy of majority voting and iMLE initialized with majority voting were both far from 1. Interestingly, in our simulations,  iMLE using SML as starting condition showed no significant improvement relative to the average balanced accuracy of SML itself.

\begin{figure}[tbph]
\begin{center}
\includegraphics[clip, trim=0cm 0.7cm 0cm 0cm, width=0.9\linewidth]{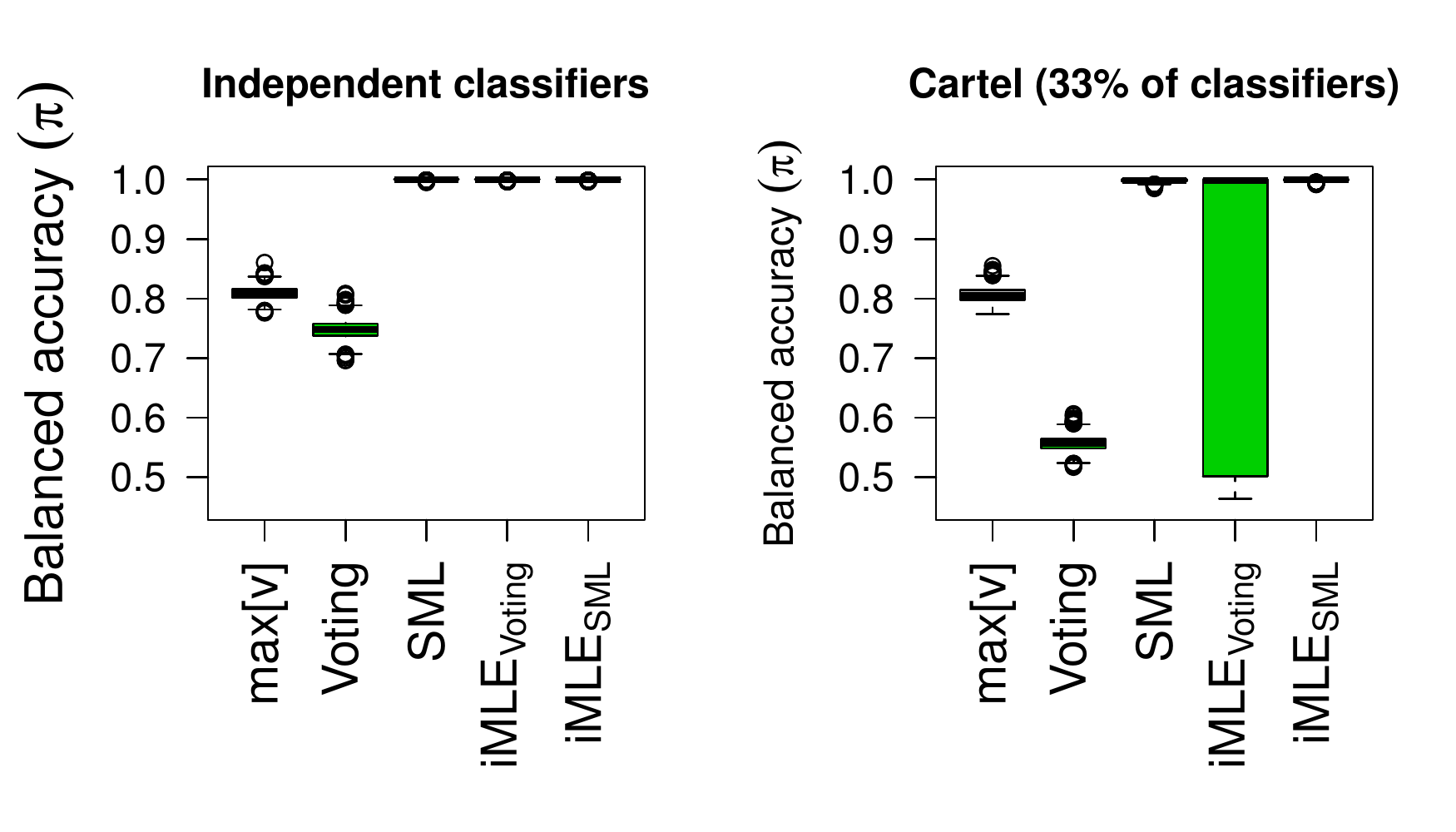}
\end{center}
\caption{Balanced accuracy of several classifiers:\ the classifier
$f_i$ with largest eigenvector entry, \(i=\argmax_j |v_j|\); majority voting; SML; and iMLE starting from SML or from majority voting. (Left Panel) 100 independent classifiers; (Right Panel) 67 honest classifiers and 33 belonging to a cartel with target balanced accuracy of 0.5. The boxplots represent the distribution of balanced accuracies of 3000 independent runs.} 
\label{MT-fig:SimPerf}
\end{figure}

\begin{figure}[tbh]
\centering
\includegraphics[clip, trim=0cm 1cm 0cm 0cm, width=\linewidth]{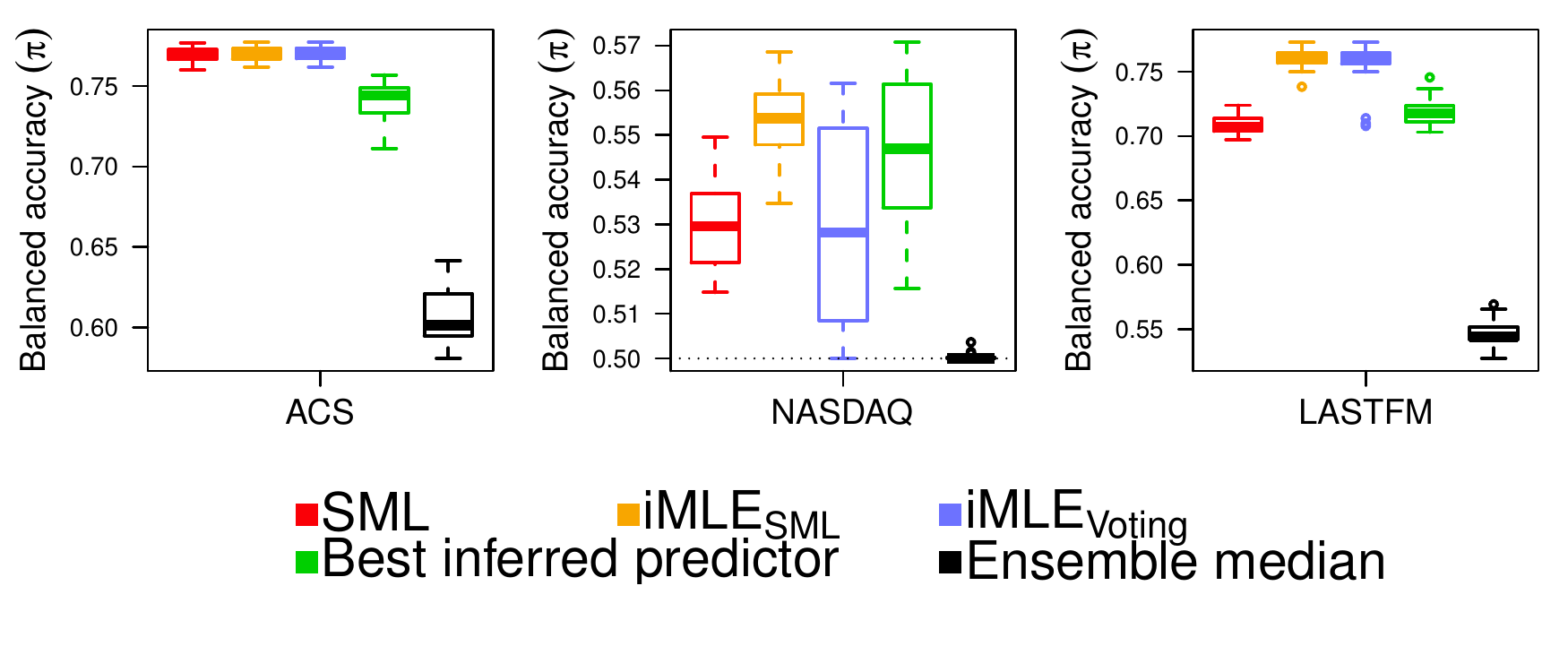}
\caption{Comparison between SML, iMLE from SML or from majority voting, the best inferred predictor and the median balanced accuracy of all ensemble predictors, on real-world datasets. The ACS dataset (left panel) approximately satisfied our assumptions. In the NASDAQ dataset (central panel) many predictors had poor performances.  For the LASTFM data (right panel), the predictors did not satisfy the conditional independence assumption. In all cases iMLE starting from SML had equal or higher balanced accuracy than iMLE starting from majority voting. The boxplots represent the distribution of balanced accuracies over 30 independent runs.} \label{MT-fig:RealData}
\end{figure}

\subsection*{Real Datasets} 
We applied our spectral approaches to 17 different datasets of moderate and large sizes from medical, biological, engineering,
financial and sociological applications. Our ensemble of predictions was comprised of 33 machine-learning methods
available in the software package Weka \cite{witten_data_2011} (see Methods). 
We split each dataset into a labeled part and an unlabeled part, the latter serving as the test data \(D\) used to evaluate our methods. To mirror our problem setting, each algorithm had access and was trained on different subsets of the labeled data (see Supplementary Information).  

Figs.~\ref{MT-fig:RealData},~\ref{SI-fig_FIG10},~\ref{SI-fig_FIG11} and ~\ref{SI-fig_FIG12} show the results of different meta-classifiers on these datasets. Let us now interpret these results and explain the apparent differences in balanced accuracy between different approaches, in light of our theoretical analysis in the previous section.  

In datasets where our assumptions are approximately satisfied, we expect SML, iMLE initialized with SML, and iMLE initialized with majority voting to exhibit similar performances. This is the case in the ACS\ data (left panel of Fig.~\ref{MT-fig:RealData}), and in all datasets in Fig.~\ref{SI-fig_FIG10}. We verified that in these datasets \eqref{MT-StatIndependence} indeed holds approximately (see Table~\ref{SI-TableCondInd}). In addition, in all these datasets, the corresponding sample covariance matrix of the 33 classifiers was almost rank-one with \(\lambda_1(\hat R)/\text{Trace}(\hat R)> 0.8$. 

Fig.~\ref{SI-fig_FIG11} and Fig.~\ref{MT-fig:RealData} (central panel) correspond to datasets where the median performance of the classifiers was only slightly above 0.5, with some classifiers having poor, even worst than random balanced accuracy. 
Interestingly, in these datasets, the covariance matrix between classifiers was far from being rank one (similar to the case when cartels were present).  
The relative amount of variance captured by the first two leading eigenvalues
$\lambda_1/ \sum \lambda_j$ and $\lambda_2/ \sum \lambda_j$
was, on average, $51.4\%$ and $15.0\%$, respectively. In these datasets, SML seems to offer a clear advantage: initializing iMLE with SML rather than with majority voting avoids the poor outcomes observed in the NYSE, AMEX and PNS datasets.

Finally, the datasets in Fig.~\ref{SI-fig_FIG12} and in Fig.~\ref{MT-fig:RealData} (right panel) 
are characterized by very sparse (ENRON) or high-dimensional (LASTFM) feature spaces $\mathcal{X}$. 
In these datasets, some instances were highly clustered in feature space, while others were isolated. 
Thus, in these datasets many classifiers made identical errors. 

Remarkably, even in these cases, iMLE initialized with SML had an equal or higher median balanced accuracy than iMLE initialized with majority voting. This was consistent across all datasets, indicating that the SML prediction provided a better starting point for iMLE than majority voting.

\section*{Summary and Discussion}
In this paper we presented a novel spectral-based statistical analysis for the problems of unsupervised ranking and combining of multiple predictors. 
Our analysis revealed that under standard independence assumptions, the off-diagonal of the classifiers covariance matrix corresponds to a rank-one matrix, whose eigenvector entries are proportional to the classifiers balanced accuracies. 
To the best of our knowledge, our work gives the first computationally efficient and asymptotically consistent solution to the classical problem posed by Dawid and Skene \cite{dawid_maximum_1979} in 1979, for which thus far only non-convex iterative likelihood maximization solutions have been proposed \cite{whitehill_whose, jin_learning_2003, lauritzen_em_1995, snow_cheap_2008, walter_estimation_1988}.

Our work not only provides a principled spectral approach for unsupervised ensemble learning (such as our SML), but also raises several interesting questions for future research. 

First, our proposed spectral-based SML has inherent limitations: it may be sub-optimal for finite samples, in particular when one classifier is significantly better than all others. Furthermore,  most of our analysis was asymptotic in the limit of an infinitely large unlabeled test set, and assuming perfect conditional independence between classifier errors. 
A theoretical study of the effects of a finite test set, and of approximate independence between classifiers on the accuracy of the leading eigenvector is of interest. This is particularly relevant in the crowdsourcing setting, where only few entries in the prediction matrix $f_i(x_k)$ are observed. While an estimated covariance matrix can be computed using the joint observations for each pair of classifiers, other  approaches that directly fit a low rank matrix may be more suitable. 

Second, a natural extension of the present work is to multi-class or regression problems where the response is categorical or continuous, instead of binary.
We expect that  in these settings the covariance matrix of independent classifiers or regressors is still approximately low-rank. Methods similar to ours may improve the quality of existing algorithms.

Third, the quality of predictions may also be improved by taking into consideration instance difficulty, discussed for example in  \cite{whitehill_whose, raykar_learning_2010}. These studies assume that some instances are harder to classify correctly, independent of the classifier employed, and propose different models for this instance difficulty. In our context, both very easy examples (on which all classifiers agree) and very difficult ones (on which classifier predictions are as a good as random) are not useful for ranking the different classifiers. Hence, modifying our approach to incorporate instance difficulty is a topic for future research. 

Finally, our work also provides insights on the effects of a malicious cartel. The study of spectral approaches to identify cartels and their target, as well as to ignore their contributions, is of interest due to its many potential applications, such as electoral committees and decision-making in trading.

\section*{Materials and methods}
\subsection*{Datasets and Classifiers}
We used 17 datasets for binary classification problems from science, engineering, data mining and finance (Table~\ref{SI-tabData}). The classifiers used are described in \cite{strino_vda_2011} or are implemented in the Weka suite \cite{witten_data_2011} (Table~\ref{SI-tabMethods}).

\subsection*{Statistical Analysis and Visualization}
Statistical analysis and visualization were performed using MATLAB (2012a, The MathWorks, Natick, MA) and R (\url{www.R-project.org}). 
Additional information is provided in the Supplementary Information.

\section*{Acknowledgments}
We thank Amit Singer, Alex Kovner, Ronald Coifman, Ronen Basri and Joseph Chang for their invaluable feedback. The Wisconsin breast cancer dataset was collected at the University of Wisconsin Hospitals by Dr. W.H. Wolberg and colleagues. F.S. is supported by the American-Italian Cancer Foundation. B.N. is supported by grants from the Israeli Science Foundation (ISF) and from Citi Foundation. Y.K is supported by the Peter T. Rowley Breast Cancer Research Projects (New York State Department of Health) and NIH (R0-1 CA158167).

\endgroup 


\newpage
\setcounter{section}{0}
\setcounter{figure}{0}
\setcounter{table}{0}
\setcounter{equation}{0}
\renewcommand{\thefigure}{S\arabic{figure}}
\renewcommand{\thetable}{S\arabic{table}}
\renewcommand{\theequation}{S\arabic{equation}}
\renewcommand{\thesection}{\Alph{section}}

\begin{center}
{\Large
\textbf{Supplementary Information: Ranking and combining multiple predictors without labeled data}\\
}
\end{center}
\begin{center}
{\large
Fabio Parisi$^{1,\#}$, 
Francesco Strino$^{1,\#}$,
Boaz Nadler$^{2}$, 
Yuval Kluger$^{1, 3, \ast}$
}
\end{center}
\footnotesize{\noindent{$^{1}$ Yale University School of Medicine, Department of Pathology, 333 Cedar St., New Haven, CT 06520, USA. $^{2}$ Weizmann Institute of Science, Department of Computer Science and Applied Mathematics, Rehovot, 76100 Israel. $^{3}$ NYU Center for Health Informatics and Bioinformatics New York University Langone Medical Center 227 East 30$^{\text{th}}$ Street, New York, NY 10016, USA.
$^{\#}$ These authors contributed equally to this work.$^{\ast}$ E-mail: yuval.kluger@yale.edu.}

\tableofcontents
\newpage

\section{Covariance between classifiers }
{\em Proof of Lemma~\ref{MT-LEMMA1}}.
 To prove the lemma we first compute the mean $\mu_i = \mathbb{E}[f_i(X)]$ and variance $Var[f_i(X)]$ of the \(i\)-th classifier. 
 We then use these results to compute the entries of the population covariance matrix,  $q_{ij}=\mathbb{E}[(f_i(X)-\mu_i) \cdot (f_j(X)-\mu_j)]$.
  
Under the assumption of independence between instances, the population mean of the  $i$-th classifier is

\[
\begin{array}{lcl}
\mathbb{E}\left[\vphantom{f_i(X)^2} f_i(X)\right] 
                        & = & \Pr[f_i(X)=1] - \Pr[f_i(X)=-1]\\ 
  & = &\sum_{y\in\{-1,1\}} \Pr[f_i(X)=1|Y=y]\Pr[Y=y] - \sum_{y\in\{-1,1\}} \Pr[f_i(X)=-1|Y=y]\Pr[Y=y] .
 \end{array}
 \]
 Using the definitions of sensitivity $\psi_i = \Pr[f_i(X)=1|Y=1]$, specificity $\eta_i = \Pr[f_i(X)=-1|Y=-1]$, and class imbalance
$b = \Pr[Y=1]-\Pr[Y=-1]$ , the equation above can be expressed as follows, 
 \begin{equation}\label{SI-eq:eq1}
  \begin{array}{lcl}
\mu_i =   \mathbb{E}\left[\vphantom{f_i(X)^2} f_i(X)\right]
 & = & \psi_{i}\left(\frac{1+b}{2}\right)+(1-\eta_{i})\left(\frac{1-b}{2}\right)-(1-\psi_{i})\left(\frac{1+b}{2}\right)-\eta{}_{i}\left(\frac{1-b}{2}\right)\\
 & = & \psi_{i}-\eta_{i}+b\left(\psi_{i}+\eta_{i}-1\right)
        = 2\delta_{i}+b(2\pi_{i} - 1)
\end{array}
\end{equation}
where $\pi_i = (\psi_i + \eta_i)/2$ is the balanced accuracy of the \(i\)-th classifier and $\delta_i = (\psi_i - \eta_i)/2$.

Similarly, the population variance of the  $i$-th classifier is
\begin{equation}
\text{\rm Var}\left[\vphantom{x^2} f_i(X)\right] = \mathbb{E}\left[f_i(X)^{2}\right]-\mathbb{E}\left[\vphantom{f_i(X)^2} f_i(X)\right]^{2} = 1-\mathbb{E}\left[\vphantom{f_i(X)^2} f_i(X)\right]^{2} = 1-\left(2\delta_{i}+b(2\pi_{i} - 1)\right)^{2}.
\end{equation}

Next, consider $\mathbb{E}[f_i(X) \cdot f_j(X)]$. 
Under the assumption of independence of errors between different instances and between different classifiers,
for \(i\neq j \)
\begin{equation}\label{SI-eq:eq3}
\begin{array}{lcll}
\mathbb{E}[f_i(X) \cdot f_j(X)] & = & \Pr[f_i(X)=f_j(X)] -\Pr[f_i(X)=-f_j(X)]\\
 & = &\left(\frac{1+b}{2}\right)\psi_{i}\psi_{j}+\left(\frac{1+b}{2}\right)(1-\psi_{i})(1-\psi_{j})
  + \left(\frac{1-b}{2}\right)(1-\eta_{i})(1-\eta_{j})+\left(\frac{1-b}{2}\right)\eta_{i}\eta_{j}\\
 &  &- \left(\frac{1+b}{2}\right)\psi_{i}(1-\psi_{j})-\left(\frac{1+b}{2}\right)(1-\psi_{i})\psi_{j}
   -  \left(\frac{1-b}{2}\right)\eta_{i}(1-\eta_{j})-\left(\frac{1-b}{2}\right)(1-\eta_{i})\eta_{j}\\
\end{array}
\end{equation}

Combining Eq.~\eqref{SI-eq:eq1} and Eq.~\eqref{SI-eq:eq3} yields that for \(i\neq j \)
\[
 \mathbb{E}[f_i(X) \cdot f_j(X)] - \mathbb{E}[f_i(X)] \cdot \mathbb{E}[f_j(X)]
=(1-b^2)(\psi_i + \eta_i - 1)(\psi_j + \eta_j - 1) 
= (1-b^2)(2\pi_i - 1)(2\pi_j - 1).
\]
Thus, the entries $q_{ij}$ of the $M \times M$ covariance matrix of the $M$ classifiers are
\begin{equation}
q_{ij} = \left\{
  \begin{array}{c c l}
    1-\mu_i^2 & &  i=j\\
    (2\pi_i-1)(2\pi_j-1)\left(1-b^2\right) & & i \neq j \\
  \end{array} \right. 
\end{equation} \hfill $\Box$.

\section{Rank-one Eigenvector Estimation}
In this section we describe four approaches to estimate the eigenvector $\mathbf{v}$ of the rank one matrix \(R\) from the sample covariance matrix $\hat{Q}$. We term these methods (i) {\bf linear system} approach; (ii) {\bf weighted linear system} approach; (iii) {\bf SDP} approach; and (iv) {\bf direct eigendecomposition} approach.
In our simulations we found that all four approaches gave comparable rankings, though the latter was slightly less accurate (Fig.~\ref{SI-fig_methodsdiagonal}).  The linear system approach (i) had computational complexity comparable to the fastest method of direct eigendecomposition, while providing a ranking of quality comparable to the much more computationally heavy SDP method. Method (i)\ was also slightly faster to compute than its weighted counterpart, method (ii), so we chose it for our benchmarks.

\begin{figure}[t]
\centering
\includegraphics[width=0.4\linewidth]{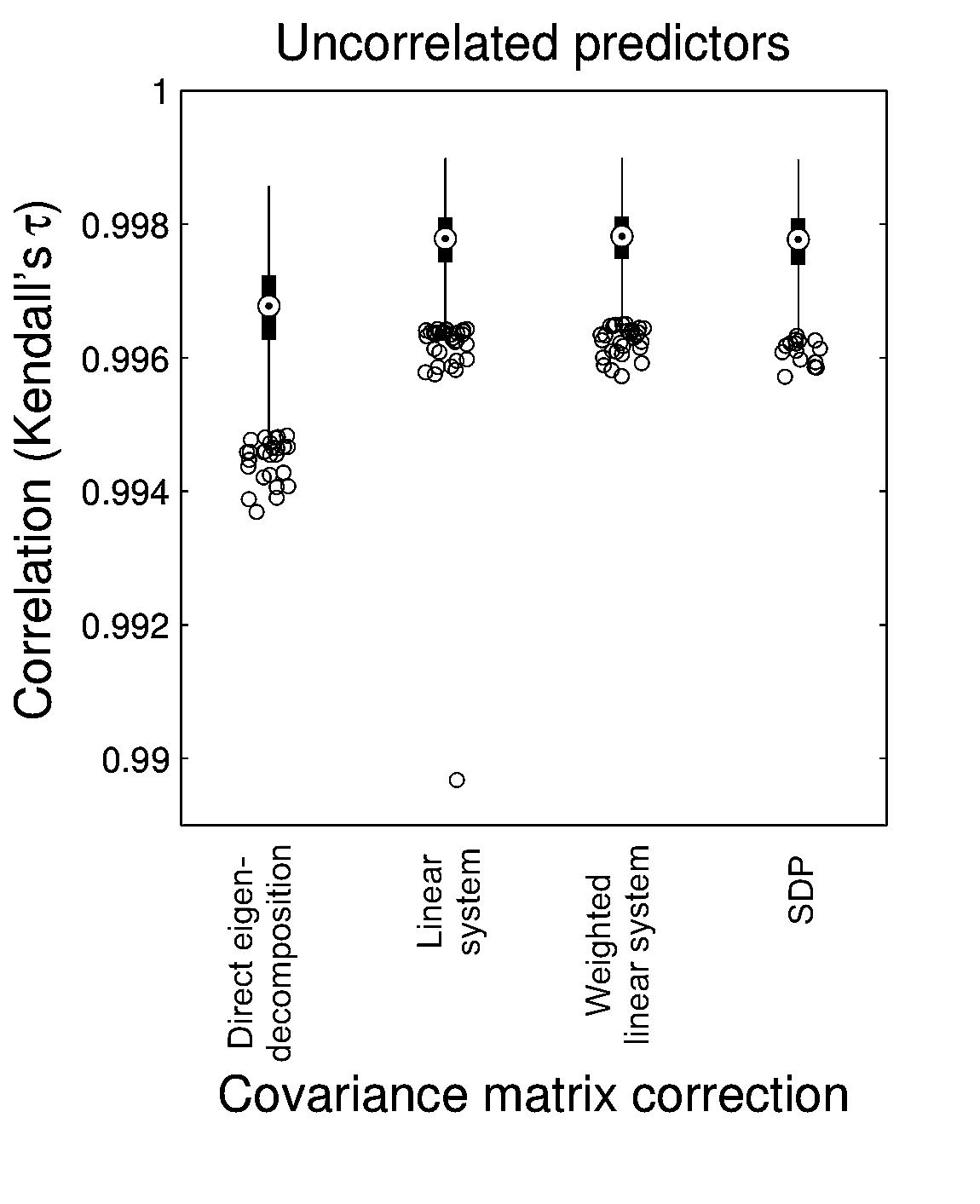}
\caption{Comparison of the four different approaches to estimate the eigenvector of the rank-one matrix $R$. The simulated data was constructed as described in section \ref{SI-sec:Simulated_Data}. The reconstruction quality is measured by  Kendall's $\tau$ correlation coefficient between the entries of the eigenvector estimated by each approach and the true eigenvector of the rank-one matrix.}
\label{SI-fig_methodsdiagonal}
\end{figure}

\subsection{Linear system}
As discussed in the main text, one approach to rank the \(M\) classifiers is to construct an estimator \(\hat R\) of the rank-one matrix \(R\), 
compute its leading eigenvector $\hat{\bf v}$ and rank the \(M\) classifiers by sorting its entries. 
Given that  \(\mathbb{E}[\hat Q] = Q\), we estimate the off-diagonal entries of \(\hat R\) simply by those of $\hat Q$, and only need a consistent method to estimate the diagonal entries.  
To this end, note that 
upon the change of variables $r_{ij} = e^{t_i} \cdot e^{t_j}$, it follows
that in the population setting,
for all \(i\neq j\), 
\[
\log |r_{ij}| = \log |q_{ij}|  - t_i - t_j = 0.
\]
In the finite sample setting, we replace the unknown \(q_{ij}\) by $\hat q_{ij}$ and look for an $M$-dimensional vector ${\bf t}$ such that the relation above  holds approximately for all pairs \(i\neq j\), 
\begin{equation}
 \mathbf{\hat{t}} = \arg\min\sum_{j>i} (\log|\hat{q}_{ij}| - 
 \hat{t}_i - \hat{t}_j)^2 .
        \label{SI-eq:hat_t}
\end{equation}
From the vector ${\bf t}$ we estimate the diagonal entries of $R$ 
as $\hat{{r}}_{ii} = \exp( {2\cdot\hat{t}_{i}})$.

As the functional in Eq.~\eqref{SI-eq:hat_t} is quadratic, the vector \(\hat {\bf t}\) is efficiently found by solving a system of linear equations with $M$ unknowns. Since  \(\hat q_{ij}\to q_{ij}\) as sample size \(S\to\infty\), it follows that \(\hat{\bf t}\) is an asymptotically consistent estimate of ${\bf t}$. Consequently the resulting \(\hat{\bf v}\) is a consistent estimate of \({\bf v}\), and asymptotically it yields a perfectly correct ranking of the \(M\) classifiers, according to their balanced accuracies.

In practice, to avoid the singularity at zero of the logarithm function, we modify Eq.~\eqref{SI-eq:hat_t} by  summing only over indices \(i,j \)  for which $|\hat{q}_{ij}| > 2\sqrt{Var[\hat{q}_{ij}]}$, where $Var[\hat{q}_{ij}]$ is a plug-in estimator of Eq.~\eqref{MT-eqvarQ} from the main text, and the factor 2 is arbitrary. 

\subsection{Weighted linear system}
Similar to the linear system approach presented above, we can instead consider the following weighted least square problem, where 
$\text{Var}[\hat{q}_{ij}]$ is given by Eq.~\eqref{MT-eqvarQ} from the main text.
\begin{equation}\label{SI-eq:WLS}
 \mathbf{\hat{t}} = 
 \arg\min\sum_{j>i} \frac{\hat{q}^2_{ij}}{\text{Var}[\hat{q}_{ij}]}\cdot (\log(|\hat{q}_{ij}|) - \hat{t}_i - \hat{t}_j)^2.
\end{equation}
The resulting estimator $\hat{\bf t}$ is also solved via a system of linear equations. 

\subsection{SDP approach}
Here we look for a  rank-one matrix \(\hat R=\hat\lambda\hat{\bf v}\hat{\bf v}^T\), whose off-diagonal terms are closest to those of \(\hat Q\). While the rank-one constraint is non-convex, its standard relaxation to a trace constraint yields 
\begin{equation}
\hat R = \arg\min\sum_{i\neq j} (\hat q_{ij}-R_{ij})^2 +\theta\, \text{Trace}(R)
        \label{SI-eq:SDP_R}  
 \end{equation}
subject to $R=R^T$, $R\succeq0$ and where $\theta$ is a suitably chosen regularization parameter. This is a convex problem, which can be solved  via semi-definite programming \cite{candes_exact_2009}. We thus term it {\bf SDP approach}.
While in principle SDP problems can be solved to arbitrary accuracy in polynomial time in $M$, this approach is significantly slower than the two previous ones, which require solutions to systems of linear equations.

\subsection{Direct eigendecomposition}
Finally, an even simpler approach is to rank the classifiers by directly computing the leading eigenvector of $\hat Q$. For a finite number of classifiers $M$, it follows from Lemma~\ref{MT-LEMMA1}  that as \(S\to\infty\), this {\bf direct eigen-decomposition approach} is generally \textit{not} consistent. However, as the following lemma shows, if the rank one matrix \(R\) has a large spectral gap, \(\lambda\gg 1\), then this leading eigenvector is close to the true one. 

\begin{lemma}\label{SI-lemma:eig_Q}
Let $\bf w$ be the leading unit-norm eigenvector of the population matrix $Q$, and let $\lambda$ be given by Eq.~\eqref{MT-eq:lambda_tilde_Q} in the main text.
Then, \begin{equation}
\left({\bf w}^T {\bf v}\right)^2 \geq 1- \frac{2}{\lambda}.
\end{equation}
\end{lemma}

{\em Proof} :
Let $\lambda(Q)$ be the leading eigenvalue of $Q$ with corresponding unit-norm eigenvector ${\bf w}$. Let $\lambda$ be the eigenvalue of the rank-one 
matrix $R$ with corresponding unit-norm eigenvector ${\bf v}$. First, note that 
\begin{equation}
Q = R +D,
        \label{SI-eq:Q_tilde_Q}
\end{equation}
where $D$ is a diagonal matrix with entries
\[
d_{ii} = 1-\mu_i^2 - (1-b^2)(2\pi_i-1)^2.
\]
Hence $\|D\|_2 = \max_i|d_{ii}| \leq 1$. It thus readily follows from Weyl's theorem that
\begin{equation}
|\lambda(Q)-\lambda|\leq \|D\|_2\leq 1.
        \label{SI-eq:lambda_Q_tilde_Q}
\end{equation}
Now, multiplying the eigenvector equation \(Q{\bf w}=\lambda(Q){\bf w}\)
from the left by ${\bf w}^T$, and inserting the relation \eqref{SI-eq:Q_tilde_Q}
gives that
\[
\lambda(Q) = \lambda \left({\bf w}^T{\bf v}\right)^2 +{\bf w}^TD{\bf w}.
\]
The lemma follows by combining Eq.~\eqref{SI-eq:lambda_Q_tilde_Q}
with the bound $|{\bf w}^T D{\bf w}|\leq 1$. \hfill $\Box$.

\hfill

Note that if all classifiers in the ensemble have a balanced accuracy
bounded away from $1/2$, then $\lambda = O(M)$ and then for \(M\gg 1,\) 
the angle between ${\bf v}$ and ${\bf w}$ is small. 

Finally, we note that this direct eigendecomposition approach is equivalent to ranking classifiers by a singular value decomposition (SVD) of the $S\times M$ mean-centered\ matrix of predicted labels $f_i(x_k)$. This approach, although apparently without the mean-centering operation, was recently suggested in \cite{karger_budget-optimal_2011_SI}, which proposed the $j$-th entry in the leading right singular vector as a proxy for the reliability of the \(j\)-th classifier. Our work provides a novel probabilistic interpretation to this approach, as it shows that the entries of \({\bf w}\),
which is also the leading right singular vector of the (mean-centered) matrix $f_i(x_k)$, are approximately those of \({\bf v}\), which in turn are proportional to the balanced accuracies of the 
classifiers.

\subsection{Asymptotic Eigenvector Stability }
\label{SI-sec:Eig_stability}

We now consider the asymptotic stability of the estimated eigenvector to small perturbations due to finite sample fluctuations in our estimate  $\hat Q$. First note that for all \(i \neq j\), $\hat q_{ij} - q_{ij} = O(1/\sqrt{S})$. It thus follows that upon solving the linear system for the vector ${\bf t}$, asymptotically its errors are also $O(1/\sqrt{S})$, and hence for all $i\neq j$, we may assume that $\hat r_{ij} - r_{ij} = O(1/\sqrt{S})$. 

To understand how these fluctuations affect the estimation of the leading eigenvector of the rank one matrix $R$, we consider the one-parameter family of matrices  \(\hat R(\epsilon) = R+\epsilon B\) where  $B=\sqrt{S}(\hat R - R)$ is a matrix whose entries are all $O(1)$. By definition, at \(\epsilon=1/\sqrt{S}\) we have that $\hat R(\epsilon)=\hat R$. We thus view $\epsilon$ as a small parameter, study the dependence of the leading eigenvector of \(\hat R(\epsilon)\) on \(\epsilon\), and eventually plug in $\epsilon=1/\sqrt{S}$. 

Given that both $R$\ and $B$ are symmetric, standard results from matrix perturbation theory \cite{Kato} imply that for sufficiently small $\epsilon$ the leading eigenvector and eigenvalue of $\hat R(\epsilon)$ are analytic functions of $\epsilon$. At $\epsilon=0$, these resort to the eigenvector ${\bf v}$ and eigenvalue $\lambda$ of the exact rank one matrix $R$. For small $\epsilon >0$ we may thus expand
\begin{eqnarray}
\hat \lambda(\epsilon) &=& \lambda+\epsilon \lambda^{(1)}+\epsilon^2\lambda^{(2)} +\ldots \nonumber \\
\hat{\bf v}(\epsilon) & = & {\bf v} +\epsilon{\bf v}^{(1)} + \epsilon^2 {\bf v}^{(2)} + \ldots \nonumber 
\end{eqnarray} 
Inserting this expansion into the eigenvalue-eigenvector equation
$\hat R(\epsilon)\hat{\bf v}(\epsilon)= \hat\lambda(\epsilon)\hat {\bf v}(\epsilon)$, and equating powers of $\epsilon$ gives that the $O(\epsilon)$ equation reads
\begin{equation}
        \label{SI-eq:O_epsilon}
R {\bf v}^{(1)}+B{\bf v} = \lambda {\bf v}^{(1)} + \lambda^{(1)}{\bf v}.
\end{equation}
Since the eigenvector $\hat{\bf v}(\epsilon)$ is defined only up to a normalization constant, we conveniently chose it to be that ${\bf v}^T\hat{\bf v}(\epsilon)=1$ for all $\epsilon$, which in particular implies that ${\bf v}^T{\bf v}^{(1)}=0$.

Now multiplying Eq.~\eqref{SI-eq:O_epsilon} from the left by \({\bf v}^T\)
gives that $\lambda^{(1)} = {\bf v}^T B{\bf v}$ and 
\begin{equation}
{\bf v}^{(1)} = \frac1{\lambda} (I-{\bf v}{\bf v}^T)^\dagger
\left(B-{\bf v}^T B{\bf v}\right) {\bf v}.
        \label{SI-eq:v_1}
\end{equation}
where $A^\dagger$ denotes the Moore-Penrose pseudo-inverse of $A$. 

The key point from Eq.~\eqref{SI-eq:v_1} is that for a given spectral gap of size $\lambda$ of the rank-one matrix $R$, asymptotically in $S$, the perturbation in the leading eigenvector estimate is $\hat{\bf v}-{\bf v} = O( \frac1\lambda \frac1{\sqrt{S}})$. 

\section{Spectral Meta-Learner}
In this section we present the derivation of the Spectral Meta-Learner (SML) as a linearization of the maximum likelihood estimator (MLE) of the vector of true class labels around $(\psi^*, \eta^*) = (1/2,1/2)$.

\subsection{Maximum Likelihood Estimator (MLE)}
Under the assumption of independence between classifier errors and between instances, given the specificities and sensitivities of the \(M\) classifiers, the overall likelihood of the labels of all \(S\) instances is a product of the likelihood of each individual instance label. Hence, for each instance $x_{k}$ its class label $y_k$ can be estimated independently of the class labels of all other instances. The MLE $\hat{y}_k^{\text{(ML)}}$ of  $y_{k}$ is
\begin{eqnarray*}
\hat{y}_k^{\text{(ML)}} & = & 
 \argmax\left\{\log\, \mathfrak{L}(f_1(x_k),\ldots,f_M(x_k);y_k=1),\log\, \mathfrak{L}(f_1(x_k),\ldots,f_M(x_k);y_k=-1)\right\}\\
 & = & \text{sign}\left( \log\, \mathfrak{L}(f_1(x_k),\ldots,f_M(x_k);y_k=1)-\log\, \mathfrak{L}(f_1(x_k),\ldots,f_M(x_k);y_k=-1)\right)\\
 & = & \text{sign}\left(
 \sum_{i|f_i(x_k)=1}\log\psi_{i}+\sum_{i|f_i(x_k)=-1}\log(1-\psi_{i})
  - 
  \Big(\sum_{i|f_i(x_k)=1}\log(1-\eta_{i}) + \sum_{i|f_i(x_k)=-1}\log\eta_{i}
  \Big)\right)\\
 & = & \text{sign}
 \Bigg(\sum_{i|f_i(x_k)=1}\left(\log\psi_{i}-\log(1-\eta_{i})\right) + \sum_{i|f_i(x_k)=-1}\left(\log(1-\psi_{i})-\log\eta_{i}\right)
 \Bigg).\\
\end{eqnarray*}

Next, note that the conditions $f_i(x_k)=1$ and $f_{i}(x_k)=-1$ in the two sums above can be represented by the following two indicator functions, 
\begin{equation*}
\frac{1+f_i(x_k)}{2} = \left\{
  \begin{array}{c c l}
    0 & &  f_i(x_k)=-1\\
    1 & &  f_i(x_k)=1 \\
  \end{array} \right.
\quad \text{and}
\qquad 
\frac{1-f_i(x_k)}{2} = \left\{
  \begin{array}{c c l}
    1 & &  f_i(x_k)=-1\\
    0 & &  f_i(x_k)=1 \\
  \end{array} \right. .
\end{equation*}
Using these indicator functions allows to express the MLE as a function of $\psi_i$ and $\eta_i$ as follows
\begin{eqnarray}
\hat{y}_k^{\text{(ML)}} & = &  \text{sign}\left(\sum_{i}\frac{1+f_i(x_k)}{2}\left(\log\psi_{i}-\log(1-\eta_{i})\right) 
        + \sum_{i}\frac{1-f_i(x_k)}{2}\left(\log(1-\psi_{i})-\log\eta_{i}\right)\right) \nonumber\\
 & = &  \text{sign}\left(\sum_{i=1}^M f_i(x_k)\log \alpha_i + \log \beta_i \right).
        \label{SI-eq:MLE_exact}
\end{eqnarray}
where
\begin{equation} 
        \label{SI-eq:alpha_beta}
\alpha_i =  \frac{\psi_{i}\eta_{i}}{(1-\psi_{i})(1-\eta_{i})}
\quad  \mbox{and}\quad \beta_i=    
\frac{\psi_{i}(1-\psi_{i})}{\eta_{i}(1-\eta_{i})}.
\end{equation}

\subsection{The SML: A first-order approximation of the MLE estimator}

Combining Eqs.~\eqref{SI-eq:MLE_exact} and \eqref{SI-eq:alpha_beta}, the maximum likelihood estimate  $\hat{y}_k^{\text{(ML)}}$ of the label \(y_{k}\) of the instance $x_k$ is 
\begin{equation}
                \label{SI-eq:hat_y_k_ML}
\hat{y}_k^{\text{(ML)}}  =  \text{sign}\left(\sum_{i} f_i(x_k) \log\left(\frac{\psi_{i}\eta_{i}}{(1-\psi_{i})(1-\eta_{i})}\right)  + \log\left(\frac{\psi_{i}(1-\psi_{i})}{\eta_{i}(1-\eta_{i})}\right) 
 \right) .
\end{equation}
A first-order Taylor expansion of the logarithms, around specificity and sensitivity values $(\psi_{i}^{*},\eta_{i}^{*})$ gives 
\begin{eqnarray*}
\sum_{i=1}^M f_i(x_k)\log \alpha_i + \log \beta_i & = & \sum_{i} f_i(x_k) \log\big(\frac{\psi_{i}^{*}\eta_{i}^{*}}{(1-\psi_{i}^{*})(1-\eta_{i}^{*})}\big)  + \log\big(\frac{\psi_{i}^{*}(1-\psi_{i}^{*})}{\eta_{i}^{*}(1-\eta_{i}^{*})}\big) \\
&  & + f_i(x_k)\left(\frac{\psi_{i}-\psi_{i}^*}{\psi_{i}^{*}} 
  + \frac{\eta_{i}-\eta_{i}^*}{\eta_{i}^{*}} 
  + \frac{\psi_{i}-\psi_{i}^*}{1-\psi_{i}^{*}} 
  + \frac{\eta_{i}-\eta_{i}^*}{1-\eta_{i}^{*}}\right)\\
&  & +\left(\frac{\psi_{i}-\psi_{i}^*}{\psi_{i}^{*}} - \frac{\eta_{i}-\eta_{i}^*}{\eta^*} - \frac{\psi_{i}-\psi_{i}^*}{1-\psi_{i}^{*}} + \frac{\eta_{i}-\eta_{i}^*}{1-\eta_{i}^{*}}\right) \\
 &  & +O((\psi_i-\psi_i^*)^2,(\eta_i-\eta_i^*)^2,(\psi_i-\psi_i^*)\cdot (\eta_i-\eta_i^*))\\
&=& \sum_{i} f_i(x_k) \log\left(\frac{\psi_{i}^{*}\eta_{i}^{*}}{(1-\psi_{i}^{*})(1-\eta_{i}^{*})}\right)  + \log\left(\frac{\psi_{i}^{*}(1-\psi_{i}^{*})}{\eta_{i}^{*}(1-\eta_{i}^{*})}\right) \\
&  & +(\psi_{i}-\psi_{i}^*) \frac{f_i(x_k) - (2 \psi_{i}^{*}-1)}{\psi_{i}^{*}(1-\psi_{i}^{*})}
+ (\eta_{i}-\eta_{i}^*) \frac{f_i(x_k) + (2 \eta_{i}^{*}-1)}{\eta_{i}^{*}(1-\eta_{i}^{*})} \\
 &  & + O((\psi_i-\psi_i^*)^2,(\eta_i-\eta_i^*)^2,(\psi_i-\psi_i^*)
  \cdot (\eta_i-\eta_i^*))
\end{eqnarray*}
At the specific values  \((\psi^*,\eta^*)=(1/2,1/2)\), where \(2\psi_i^*-1=2\eta_i^*-1=0\), the Taylor expansion above simplifies considerably. Inserting the resulting expression back into Eq.~\eqref{SI-eq:hat_y_k_ML} yields
\begin{equation*}
\hat{y}_k^{\text{(SML)}} = \text{sign}\left(\sum_{i} f_i(x_k) \left(\psi_{i} + \eta_{i} - 1\right)\right) = \text{sign}\left(\sum_{i} f_i(x_k) \left(2 \pi_{i} - 1\right)\right) = \text{sign}\left(\sum_{i} f_i(x_k) v_i\right),
\end{equation*}
where  ${\bf v}\in \mathbb{R}^M$ is the leading eigenvector of the rank-one matrix \(R\), as described in the main text. We thus call this novel ensemble-classifier
the {\em Spectral Meta-Learner} (SML).

\section{Comparison between SML and Majority Voting}
In the present section we provide insights into the potential advantages of SML over majority voting. To this end, we study the performance of these two unsupervised ensemble learners in the specific case where all classifiers, except one, have equal sensitivities and specificities. We prove that the resulting balanced accuracy of the weighted voting scheme employed by SML is greater than or equal to the balanced accuracy of majority voting. It is also greater than the balanced accuracy of the best algorithm in the ensemble, up to a small constant. 

\begin{lemma}\label{SI-lemma:SML_Vs_Voting}
Consider an ensemble of \(M\) conditionally independent classifiers such that the first classifier has sensitivity and specificity \(\psi_1=\eta_1=\pi_1\), and the remaining \(M-1\) classifiers have all the same specificity and sensitivity \(\psi=\eta=\pi$. Let $\pi_\text{Vo}$ be the resulting balanced accuracy of majority voting and  let $\pi_\text{SML}$ be the balanced accuracy of an (oracle) SML classifier, whose weights assume perfect knowledge of the values $\pi_1$ and $\pi$. Then,
for any value of \(\pi_1\) and $\pi$ (with $\pi>1/2$), 

(i) The balanced accuracy of SML is always greater than or equal to that of majority voting, 
\begin{equation}
\pi_\text{SML} \geq \pi_\text{Vo}.
\end{equation}

(ii) The balanced accuracy of SML is always greater than or equal to that of the \(M-1\) classifiers,
\begin{equation}
\pi_\text{SML} \geq \pi.
\end{equation}

(iii) The balanced accuracy of SML is greater than or equal to that of the first classifier, up to a small constant,
\begin{equation}
\pi_\text{SML} \geq \pi_1 -  \exp\left(-2\epsilon^2 (M-1) \right),
        \label{SI-eq:SML_vs_pi1}
\end{equation}

with $\epsilon = \left( 1 + (M-1)(2\psi-1)^2\right)/\left( 2(M-1)(2\psi-1)\right)$.

\end{lemma}

\hfill\\
{\bf Remarks:} Albeit for the specific case where $\pi_1=\pi$, this lemma yields five insights: 

(i) The performance of SML is higher than that of majority voting. Intuitively, this is expected since SML, being a Taylor approximation of the MLE, has weights closer to the optimal ones, in contrast to the equal weights employed by majority voting. 

(ii) The second insight is that SML is more accurate than most classifiers in the ensemble. This is not necessarily true for majority voting. For example, in a challenging classification problem where most classifiers in an ensemble are slightly better than random and one classifier is much worse than random, majority voting can have a balanced 
accuracy smaller than 1/2. 

(iii) Eq.~\eqref{SI-eq:SML_vs_pi1} may seem disappointing at first sight, as it states that there may be cases where SML has a lower accuracy than the best classifier in the ensemble. However, this is to be expected, since SML follows from a Taylor expansion of the maximum likelihood solution at specificity and sensitivity values of 1/2 (e.g., close to being totally random). Thus, SML is a 
{\em conservative} meta-classifier.   For example, if the first classifier had perfect balanced accuracy, $\pi_1=1$, then its weight in the maximum likelihood solution would be infinite,
with effectively zero weights for all other classifiers, see Eq.~\eqref{SI-eq:alpha_beta}. In contrast, SML gives finite and non-zero weights  to all classifiers, provided they are not totally random ($\pi\neq 1/2$). Hence, it may in general be worse than the best classifier in the ensemble. Eq.~\eqref{SI-eq:SML_vs_pi1} states, however, that even in this extreme case, the difference in performance between SML and the best classifier is small and it decreases exponentially with the number of classifiers. 

(iv) For simplicity we state and prove the lemma assuming that the exact values of \(\pi\) and $\pi_1$ are provided by an oracle. As discussed in Section \ref{SI-sec:Eig_stability}, with a finite unlabeled dataset consisting of \(S\) samples, these values can be estimated with accuracy $O(1/\sqrt{S})$. These estimation errors affect only the SML classifier (as majority voting gives equal weights to all classifiers), and imply that claims (i), (ii) and (iii) hold, up to additional small $O(1/\sqrt{S})$ terms. 

(v) Both in the statement of the lemma and in its proof, when a weighted ensemble classifier of the form \(\text{sign}(\sum_j a_j f_j(x)\)) gives a result of zero for the argument inside the sign, to output a $\pm 1$ class label, we flip a coin at random with probability 1/2 and output its result.

\hfill

{\em Proof:} Under the assumptions of the lemma, it follows that for the corresponding majority voting classifier $\pi_\text{Vo} = \psi_\text{Vo} = \eta_\text{Vo}$, and similarly, $\pi_\text{SML} = \psi_\text{SML} = \eta_\text{SML}$. Hence, it suffices to show that claims (i), (ii) and (iii) hold only for the respective sensitivities. 

We start by proving claim (i). To this end, we first consider the case where \(\pi_1=\pi\), or equivalently $\psi_1=\psi$. In this case SML and majority voting yield the same classifier, whose sensitivity is given by the probability that more than half of the classifiers make a correct prediction. This probability is given by the tail of the binomial cumulative distribution function, 
\begin{equation}  \label{SI-eq:votperf1}
 \psi_{\text{SML}}\Big|_{\psi_1 = \psi} = 
 \psi_{\text{Vo}} \Big|_{\psi_1 = \psi} = 
 \psi_\text{equal} = \sum_{j>\lfloor M/2 \rfloor}^{j \leq M} \psi^j (1-\psi)^{M-j}{\binom{M}{j}} = 1 - F\left(\frac{M}{2};M,\psi\right),
\end{equation}
where $\lfloor M/2 \rfloor$ denotes the floor (or integer truncation) operation and $F(k;n,p)$ is the probability of at most  $\lfloor k\rfloor$ successes in a Binomial distribution with $n$ independent trials of success probability $p$,
\begin{equation}\label{SI-eq:binomialcdf}
F(k;n,p) = \sum_{i=0}^{\lfloor k \rfloor} {\binom{n}{i}} p^i (1-p)^{n-i}.
\end{equation} 

Next, we analyze the sensitivity of majority voting when $\psi_1\neq \psi$. By conditioning on the outcome of the first algorithm (giving either a correct or incorrect prediction), it follows from Eq.~\eqref{SI-eq:votperf1} that
\begin{eqnarray}
\psi_\text{Vo} &=&  \psi_1\left[1-F\left(\tfrac{M}{2} - 1;M-1,\psi\right)\right] + (1-\psi_1)\left[1-F\left(\tfrac{M}{2};M-1,\psi\right)\right] \nonumber\\ 
  & = & 
    1 - F\left(\tfrac{M}{2};M-1,\psi\right) + \psi_1 \left[F\left(\tfrac{M}{2};M-1,\psi\right)-F\left(\tfrac{M}{2} - 1;M-1,\psi\right)\right].
\end{eqnarray}
Importantly,  $\psi_\text{Vo}$ depends linearly on $\psi_1$ 
and thus its partial derivative $\partial \psi_\text{Vo}/\partial \psi_1$
is constant
\begin{equation}
                \label{SI-eq:votingderivative}
\frac{\partial \psi_\text{Vo}}{\partial \psi_1} = F\left(\tfrac{M}{2};M-1,\psi\right)-F\left(\tfrac{M}{2} - 1;M-1,\psi\right)
      =F(\tfrac{M-1}2+\tfrac12; M-1,\psi)-F(\tfrac{M-1}2-\tfrac12; M-1,\psi).
\end{equation}

Now we consider the sensitivity of the SML classifier. Recall that in SML the first classifier is weighted differently from the other classifiers in the ensemble, proportionally to $2\psi_1 - 1$. We define its {\bf relative weight} as 
\begin{equation}
        \label{SI-eq:theta}
\theta = \frac{2\psi_1 - 1}{2\psi-1}.
\end{equation}
Again conditioning on the outcome of the first classifier, we have that
\begin{eqnarray}
\psi_\text{SML} = \psi_1\left[1-F(\tfrac{M-1}2-\tfrac\theta 2;M-1,\psi)\right]
+ (1-\psi_1)\left[1-F(\tfrac{M-1}2+\tfrac\theta 2;M-1,\psi)\right].
\end{eqnarray}

Note that due to the floor operation in computing the stair-case cumulative distribution function \(F\), $\psi_\text{SML}$ is a piecewise linear function of $\psi_1$. It is thus not differentiable at values of $\psi_1$ for which $(M-1)/2 \pm \theta/2$ is an integer. In addition, some values of $\psi_1$ for which $(M-1)/2 \pm \theta/2$ is an integer correspond to isolated local minima in the function of $\psi_\text{SML}$. These local minima can be effectively replaced by their left (or right) limit $\lim_{\tilde{\theta} \rightarrow \theta_\pm} \psi_\text{SML}$, thus obtaining a piecewise linear function without isolated points.

At any other value of $\psi_1$, $\psi_\text{SML}$ can be differentiated w.r.t. $\psi_1$. 
Since the cumulative distribution is constant for sufficiently small positive or negative changes in $\psi_1$, it follows that
\begin{equation}
        \label{SI-eq:derivative_psi_SML}
\frac{\partial \psi_\text{SML}}{\partial \psi_1} =F(\tfrac{M-1}2+\tfrac\theta 2;M-1,\psi) -F(\tfrac{M-1}2-\tfrac\theta 2;M-1,\psi). 
\end{equation}
Comparing Eq.~\eqref{SI-eq:derivative_psi_SML} to Eq.~\eqref{SI-eq:votingderivative}, we note that for $\psi_1 > \psi$, for which $\theta > 1$, we have that 
$\frac{\partial \psi_\text{SML}}{\partial \psi_1} \geq \frac{\partial \psi_\text{Vo}}{\partial \psi_1}$, whereas for $\psi_1 < \psi$, for which $\theta<1,$ it follows that  $\frac{\partial \psi_\text{SML}}{\partial \psi_1} \leq \frac{\partial \psi_\text{Vo}}{\partial \psi_1}$.
Since at $\psi_1=\psi$, the two ensemble classifiers coincide, it follows that claim (i) holds (see Fig.~\ref{SI-fig:lemmaExample} for an illustrative example).

Now we turn to prove claim (ii), that $\psi_{SML}\geq \psi$. To this end, note that when the first algorithm is random, $\pi_1=\psi_1=1/2$, according to Eq.~\eqref{SI-eq:theta} we have $\theta=0$, and thus, from Eq.~\eqref{SI-eq:derivative_psi_SML} 
it follows that 
\[
\frac{\partial \psi_\text{SML}}{\partial \psi_1}\Big|_{\psi_1=1/2} = 0.
\]
Furthermore, for $\psi_1 >1/2$ this derivative is positive, whereas for $\psi_1<1/2$ the derivative is negative. We thus conclude that $\psi_1=1/2$ is a \textit{global minima} of $\psi_\text{SML}$ as a function of $\psi_1$. Furthermore, when $\psi_1=1/2$, the first algorithm has no weight, and the sensitivity of the SML classifier is the same as that of majority voting, based on $M-1$ conditionally independent classifiers, all with balanced accuracy equal to $\psi$. It thus readily follows that claim (ii) holds. 

To finish the proof of the lemma, we now consider claim (iii). First, observe that if $\psi > \psi_1$, then $\psi_\text{SML} > \psi_1$. We therefore focus on the case $\psi_1 > \psi$. When \(\psi_1=1\), $\theta=1/(2\psi-1)$ and 
\begin{equation}
\psi_\text{SML}\Big|_{\psi_1=1} = 1-
F\left(\tfrac{M-1}2 - \tfrac12\tfrac{1}{2\psi-1};M-1,\psi\right).
\end{equation}
As discussed in remark (iii) after the lemma, the value of $\psi_\text{SML}$ can be strictly smaller than one in this case, meaning that SML is not always as good as the best classifier in the ensemble. 

However, we now show that if the \(M-1\) remaining classifiers have balanced accuracy better than random, then SML has balanced accuracy close to 
\(\pi_1\). To prove Eq.~\eqref{SI-eq:SML_vs_pi1} of claim (iii), first note that according to Eq.~\eqref{SI-eq:derivative_psi_SML}, $\partial \psi_\text{SML}/\partial \psi_1 \in [0,1]$ for all $\psi_1 \geq \psi$. Hence, as $\psi_1$ is decreased from a value of 1, $\psi_\text{SML}$ decreases {\em slower} than $\psi_1$ itself. Thus, to prove the claim, it suffices to show that at the extreme case $\psi_1=1$,
\[
F\left(\tfrac{M-1}2 - \tfrac12\tfrac{1}{2\psi-1};M-1,\psi\right) \leq \exp(-2\epsilon^2(M-1)).
\]

To this end, we apply Hoeffding's inequality for i.i.d. Bernoulli random variables (namely that for a random variable $X\sim Bin(n,\psi)$, $\Pr[X\leq n(\psi-\epsilon)] = F(n(\psi-\epsilon); n,\psi) \leq \exp(-2\epsilon^2 n))$. In our case, \(n=M-1,\) and comparing
\begin{equation}
\frac{M-1}{2} - \frac{1}{2}\frac{1}{2\psi-1} = (M-1)(\psi - \epsilon).
\end{equation}
gives $\epsilon = \left( 1 + (M-1)(2\psi-1)^2\right)/\left( 2(M-1)(2\psi-1)\right)$.
Plugging this into Hoeffding's inequality concludes the proof. 
\hfill$\Box$.

\section{Covariance between classifiers in presence of a cartel}
{\em Proof of Lemma~\ref{MT-LEMMA3}}.
As in the proof of Lemma~\ref{MT-LEMMA1}, for each classifier \(f_{i}\)
 we first compute its mean and variance, $\mu_i = \mathbb{E}[f_i(X)]$ and $Var[f_i(X)]$, respectively.
 We then use these results to compute the entries of the population covariance matrix, 
 $q_{ij}=\mathbb{E}[(f_i(X)-\mu_i)\cdot (f_j(X)-\mu_j)]$.
  
The mean and variance of honest classifiers with indices $i \in P$ have already been computed in the proof of Lemma~\ref{MT-LEMMA1}. We now consider the mean and variance of classifiers $i \in C$ that belong to the cartel. For brevity, we denote by \(\psi_c,\eta_c\) and $\pi_c$ the specificity, sensitivity and balanced accuracy of the cartel target with respect to the ground truth, 
\begin{equation*}
\psi_c = \Pr[T=1|Y=1], \quad
\eta_c = \Pr[T=-1|Y=-1],\quad
\pi_c = \tfrac12(\psi_c + \eta_c).
\end{equation*}
Furthermore, for each $i\in C$, we denote by $p_i$ and $n_i$ its specificity and sensitivity w.r.t. the cartel target,
\begin{equation}
p_i = \Pr[f_i(X)=1|T=1], \quad \quad
n_i = \Pr[f_i(X)=-1|T=-1].
\end{equation}

Under the assumption of independence between instances, the mean of a cartel member with $i \in C$ is
\[
\begin{array}{lcl}
\mathbb{E}\left[\vphantom{f_i(X)^2} f_i(X)\right] 
                        & = & \Pr[f_i(X)=1] - \Pr[f_i(X)=-1]\\
 & = & \sum_{t,y\in\{-1,1\}}\Pr[f_i(X)=1|T=t]\Pr[T=t|Y=y]\Pr[Y=y] \\
 &   & - \sum_{t,y\in\{-1,1\}}\Pr[f_i(X)=-1|T=t]\Pr[T=t|Y=y]\Pr[Y=y]
\end{array}
\]

which, after simple algebraic manipulations, simplifies to
\begin{equation}
\mathbb{E}\left[\vphantom{f_i(X)^2} f_i(X)\right]  = b(1-\psi_c -\eta_c +n_i(\psi_c+\eta_c -1) + p_i(\psi_c+\eta_c-1)) +n_i(\psi_c-\eta_c-1) + p_i(\psi_c-\eta_c+1) +\eta_c - \psi_c.
\end{equation}
Similarly, as in Lemma~\ref{MT-LEMMA1}, the population variance of the  $i$-th classifier is
\begin{equation}
\text{\rm Var}\left[\vphantom{x^2} f_i(X)\right] = \mathbb{E}\left[f_i(X)^{2}\right]-\mathbb{E}\left[\vphantom{f_i(X)^2} f_i(X)\right]^{2} = 
1-\mathbb{E}\left[\vphantom{f_i(X)^2} f_i(X)\right]^{2}.
\end{equation} 

Next, we compute $\mathbb{E}[f_i(X) \cdot f_j(X)]$. The case $i,j \in P$ was already considered in the proof of Lemma~\ref{MT-LEMMA1}, whereas the case $i,j \in C$ can be deduced from it, with the truth replaced by the cartel's target $T$. Thus,  
\begin{equation}
\mathbb{E}[f_i(X) \cdot f_j(X)] = 
\left\{
\begin{array}{cl}
(2\pi_i-1)(2\pi_j-1)(1-b^2) & i\neq j,\, i \in P, j \in P \\
(2\xi_i-1)(2\xi_j-1)(1-b^2) & i\neq j,\, i \in C, j \in C
\end{array}
\right.
\end{equation}

It thus remains to compute $\mathbb{E}[f_i(X) \cdot f_j(X)]$ for the mixed case with $i \in P$ and $j \in C$. Under the assumption of independence of errors between different instances and between different classifiers, 
\begin{equation*}
\begin{array}{lcll}
\mathbb{E}[f_i(X) \cdot f_j(X)] & = & \Pr[f_i(X)=f_j(X)] -\Pr[f_i(X)=-f_j(X)]\\
 & = & ((2\psi_i-1)((1-2n_j)(1-\psi_c)-(1-2p_j)\psi_c))(1+b)/2 \\
&& +((2\eta_i-1)((1-2p_j)(1-\eta_c)-(1-2n_j)f))(1-b)/2.
\end{array}
\end{equation*}
Combining the three equations above yields that for $ i \in P, j \in C$
\begin{equation*}
\begin{array}{lcl}
 \mathbb{E}[f_i(X) \cdot f_j(X)] - \mathbb{E}[f_i(X)] \cdot \mathbb{E}[f_j(X)] & = & (1-b^2)(\psi_i + \eta_i - 1)(\psi_c + \eta_c - 1)(n_j + p_j - 1) \\
 & = & (1-b^2)(2\pi_i - 1)(2\pi_c - 1)(2\xi_j -1).
\end{array}
\end{equation*}

Thus, the entries $q_{ij}$ of the $M \times M$ covariance matrix between the $M$ classifiers are
\begin{equation}
        \label{SI-eq:Q_rank2}
q_{ij} = \left\{
  \begin{array}{c c l}
    1-\mu_i^2 & &  i=j\\
    (2\pi_i-1)(2\pi_j-1)\left(1-b^2\right) & & i\neq j,i\in P, j\in P\\
    (2\pi_i-1)(2\pi_c-1)(2\xi_j-1)\left(1-b^2\right) & & i\in P, j\in C\\
    (2\xi_i-1)(2\xi_j-1)\left(1-b^2\right) & & i\neq j,i\in C, j\in C
  \end{array} \right. 
\end{equation} \hfill $\Box$.

\section{Matrix rank and leading eigenvectors in presence of a cartel}

{\em Proof of Theorem~\ref{MT-THEO1}}.
To simplify notation, we make the following convenient change of variables:
\[
\rho_i = 2\pi_i -1, \quad 
\tau_i = 2\xi_i - 1,
\quad 
u = (1-b^2),
\quad
\mbox{and}
\quad
\rho_c=2\pi_c-1,
\]
where $\pi_c$ is the balanced accuracy of the cartel with respect to the truth. In this notation, for 
indices $i\in P,j\in C$ as an example, we have the compact representation $q_{ij} = u\rho_i\tau_j \rho_c$.  

Our proof of the theorem is constructive: we explicitly construct \(\lambda_1,\lambda_2\in\mathbb{R}\) and two orthonormal vectors ${\bf e}_1,{\bf e}_2\in\mathbb{R}^M$ such that for all \(i\neq j\)\ 
\begin{equation}\label{SI-Ansatz0}
 q_{ij} = \lambda_1 e_{1i} e_{1j} + \lambda_2 e_{2i} e_{2j}.
\end{equation}
Furthermore, as we prove below, these eigenvectors have in fact the following specific form
\begin{equation}\label{SI-Ansatz1.5}
\sqrt{\lambda_1} e_{1i}  =  \left\{
 \begin{array}{r c l}
  \sqrt{u} \cdot a_{11} \rho_i & & i\in P\\
  \sqrt{u} \cdot a_{21} \tau_i & & i\in C\\
 \end{array} \right.
\quad\quad
\sqrt{\lambda_2} e_{2i}  = \left\{
 \begin{array}{r c l}
  \sqrt{u} \cdot a_{12} \rho_i & & i\in P\\
  \sqrt{u} \cdot a_{22} \tau_i & & i\in C\\
 \end{array} \right.
\end{equation}
where $a_{11},a_{12},a_{21},a_{22}$ are scalars yet to be determined. 

The requirement that the eigenvectors \({\bf e}_1,{\bf e}_2\) are orthogonal, namely that \(\sum_i e_{1i}e_{2i}=0\), implies that

\begin{equation}\label{SI-orthogonality}
 a_{11} a_{12} \sum_{i \in P} \rho_i^2  + a_{21} a_{22} \sum_{j \in C} 
 \tau_j^2= 0.
\end{equation}
Next, 
comparing the exact values of \(q_{ij}\), Eq.~\eqref{SI-eq:Q_rank2}, with our assumed form above gives the following set of equations, 
\begin{equation}\label{SI-Ansatz1}
\left\{
  \begin{array}{c c l}
    u \rho_i \rho_j = u \cdot a_{11}^2 \rho_i \rho_j  + u \cdot a_{12}^2 \rho_i \rho_j  &&  i\in P, j\in P\\\
    u \rho_i \rho_c \tau_j = u \cdot a_{11} a_{21} \rho_i \tau_j  + u \cdot a_{12} \rho_i \cdot a_{22} \tau_j  &&  i\in P, j\in C\\\
    u \tau_i \tau_j  = u \cdot a_{21} \tau_i \cdot a_{21} \tau_j  + u \cdot a_{22} \tau_i \cdot a_{22} \tau_j && i\in C, j\in C
  \end{array} \right.
\end{equation}
Hence, for Eq.~\eqref{SI-Ansatz0} to hold, $a_{11}$, $a_{12}$, $a_{21}$ 
and $a_{22}$ 
should satisfy the following set of equations 
\begin{equation}\label{SI-Ansatz2}
\left\{
  \begin{array}{r c l}
    a_{11}^2 + a_{12}^2 &=&1 \\
    a_{11} a_{21} + a_{12}a_{22} &=& \rho_c \\
    a_{21}^2 + a_{22}^2 &=&1 \\
    (a_{11}a_{12})/(a_{21}a_{22}) &=& -\big(\sum_{j \in C} \tau^2_j \big)/\big(\sum_{i \in P} \rho^2_i \big)=-\lambda_C/\lambda_P
  \end{array} \right. , 
\end{equation}
where $\lambda_C = (1-b^2)\sum_{i\in C}(2\xi_i-1)^2$ and $\lambda_P=(1-b^2)
\sum_{i\in P} (2\pi_i-1)^2$. 

We now show that this set of equations indeed has a unique solution, up to the trivial sign ambiguities in the definition of the two eigenvectors.  To this end, note that the following change of variables,  $a_{11} = \cos\alpha, a_{12}=\sin\alpha, a_{21}=\sin\beta,$ and $a_{22}=\cos\beta$, 
reduces the system in Eq.~\eqref{SI-Ansatz2} to

\begin{equation}\label{SI-Ansatz4}
\left\{
  \begin{array}{r c l}
    \sin\left(\alpha+\beta\right) &=& k_1 \\
    \sin\left(2\alpha\right)/\sin\left(2\beta\right) &=& -k_2
  \end{array} \right. 
\end{equation}
where $k_1=\rho_c$ and $k_2=\lambda_C/\lambda_P$. 

To solve this system, note that standard trigonometric equalities applied to the first equation above give that
\begin{equation}
        \label{SI-eq:double_angle}
\sin2(\alpha+\beta) = 2k_1\sqrt{1-k_1^2}\quad \text{and}
\quad
\cos 2(\alpha+\beta) = 1-2k_1^2.
\end{equation}
Next, rewrite the second equation as $\sin(2(\alpha+\beta)-2\beta)+k_2\sin(2\beta)=0$, and expand the first term. This gives
\[
\sin(2\alpha) + k_2\sin2(\alpha+\beta)\cos(2\alpha) - k_2\cos(2(\alpha+\beta))\sin(2\alpha) = 0
\]
or equivalently, 
\[
\tan(2\alpha) = -\frac{k_2 \sin(2(\alpha+\beta))}{1-k_2\cos(2(\alpha+\beta))}.
\]
Combining this with Eq. \eqref{SI-eq:double_angle} gives
\[
\alpha = \frac{1}{2}\arctan\left(\frac{k_1k_2}{k_2(1-2k_1^2)-1}\right).
\]

Similarly, writing the second equation  as 
$\sin(2(\alpha+\beta) - 2\beta) + k_2\sin(2\beta)=0$ and expanding gives
\[
\tan(2\beta) = - \frac{\sin(2\delta)}{k_2-\cos(2\delta)} 
\]
whose solution is 
\[
\beta = \frac{1}{2}\arctan\left(\frac{2k_1\sqrt{1-k_1^2}}{1 - k_2 - 2k_1^2}\right).
\]

Consistent with the sign ambiguity of the eigenvectors, these solutions for $\alpha$ and $\beta$ are unique up to a rotation with periodicity $\frac{\pi}{2}$. The expressions for the eigenvectors and their respective eigenvalues readily follow by back-substitution into Eq.~\eqref{SI-Ansatz1.5}.
\hfill$\Box$

\section{Simulations and benchmarks}
The following section describes how we generated the simulated data and how we performed the benchmarks. For each component of the simulation we also provide pseudo-code.

\subsection{Simulated data: Ensembles of statistically independent predictions}
\label{SI-sec:Simulated_Data}
We generated ensembles of statistically independent predictions using the random detector with fixed balanced accuracy (RDFBA) algorithm \cite{REFVDA}.  A generic RDFBA predictor with pre-determined empirical balanced accuracy  $\pi$ on a 
test set with $T$ samples,  is denoted as RDFBA($\pi$). Given a test data with \(T\) samples, a collection of RDFBAs is constructed such that any two classifiers are conditionally independent and such that their {\it empirical} balanced accuracy on the test data is equal to $\pi$. Note that two RDFBAs with the same balanced accuracy \(\pi\) may nonetheless have different sensitivity $\psi$ and specificity $\eta$. 

To briefly describe the construction we use the following standard notation: Let $\text{P}$ be the number of positives, i.e. the number of instances whose true class label is +1; $\text{N}$ is the number of negatives, where $T=\text{P}+\text{N}$; $\text{FP}$ is the number of false positives, i.e. the number of negatives that have been mistakenly predicted as positives; $\text{FN}$ is the number of false negatives. An RDFBA($\pi$) classifier is constructed from the ground truth vector $y$ as follows:
\begin{enumerate}
\item Initialize the entries of the prediction vector $f(x)$ with the corresponding entries in the ground truth ${\bf y}$.
\item Under the constraint that $\text{FN} = (2 - 2\pi - \text{FP}/\text{N})\cdot \text{P}$ is an integer, draw a random integer  $\text{FP}$ with uniform probability from $[0,\text{N}]$. 
\item $\text{FP}$ randomly chosen instances in $f(x)$, whose true label is $-1$, are assigned the wrong class label, $+1$. 
\item $\text{FN}$ randomly chosen instances in $f(x)$, whose true label is $+1$, are assigned the wrong class label, $-1$.
\end{enumerate}

In our simulations, we used $\pi \sim \mathbb{U}(0.3,0.8)$ and a total of $T = 10000$ samples, from which we randomly sampled 300 positive and 300 negative instances, to form our test data $D$ of size $S=600$ samples. Hence, the empirical balanced accuracy of the RDFBA classifiers on the test data $D$ may be slighly higher than 0.8 or lower than 0.3. 

\subsection{Simulated data: Ensembles of independent predictors with one cartel present}
To generate datasets of conditionally independent predictors which include a cartel with  $r\cdot M$ predictors, we applied the following steps: First, we generated an ensemble $P$ of $(1-r)M$ independent predictions as described above for the ground truth vector $\mathbf{y}$. Then, using another RDFBA predictor, we constructed the cartel's target vector $\mathbf{c}$, such that it had an empirical balanced accuracy $\pi_c$ with respect to the ground truth. Next, using this vector $\bf c$ we constructed an ensemble $C$ of independent predictions, as in the procedure described above, with the only difference that the balanced accuracies of all members of the cartel relative to the cartel's target were set to be equal to $0.7$.
The dataset is obtained by the union of the two ensembles of predictions, $P$ and $C$.
In our simulations we used $\pi_c = 0.5$ thus obtaining a cartel's target that is orthogonal to the ground truth.

\subsection{Real data: Ensembles of predictions from standard machine-learning classifiers}
To generate ensemble of predictions from standard machine-learning classifiers on real data, we trained the classifiers on partially overlapping training data and collected their predictions obtained on the same test data, which was independent from all the training data. 
In detail, from each dataset we sampled 600 instances (or all the instances if less than 600 were available), half of which (up to 300) were used for testing. Independently for each classifier, we selected a random subset comprising of 90\% of the instances reserved for training and used this subset as a "private" training set. The purpose of this procedure was to produce training data that was slightly different between the different classifiers, while at the same time allowing to have a significantly large number of training samples even in the smaller datasets. We chose to use at most 600 instances to reduce computational time. 
To determine the empirical distribution of performances of each classifier and of the ensemble approaches discussed in the manuscript, for each dataset we repeated this procedure 1000 times, unless otherwise specified in the figure caption.

\subsection{Custom Real datasets: Ensembles of predictions from standard machine-learning classifiers}
To generate ensemble of predictions from standard machine-learning classifiers on custom datasets from big-data repositories, we trained the classifiers on non-overlapping training data and collected their predictions obtained on the same testing data, which was independent from all the training data. 
In detail, from each dataset we sampled 50,000 instances (or half of the instances if less than 50,000 were available), and, independently for each classifier, we selected a random subset comprising of 500 instances for training. The purpose of this procedure was to produce training data that had the potential to be markedly different between the different classifiers. We chose to use at most 500 instances to reduce the computational time and memory usage required for training. 
To determine the empirical distribution of performances of each classifier and of the ensemble approaches discussed in the manuscript, for each dataset we repeated this procedure 30 times, unless otherwise specified in the figure caption.

\section{Custom datasets}
In addition to eight standard machine learning datasets from the UCI repository, which are described in the first part of Table \ref{SI-tabData}, we created nine additional datasets from publicly available data in the fields of economics, sociology, geography, semantics, ecology, and finance (see second part of this table).

As these datasets are not readily available, we provide scripts to generate the corresponding matrices of features and class labels. These matrices can be used to train the set of 33 standard machine learning algorithms described in Table~\ref{SI-tabMethods} and subsequently apply the SML and iMLE approaches described in the main text. 

The scripts are available at \url{http://sourceforge.net/projects/klugerlab/files/SML_customdatasets}

\subsection{ACS}
This dataset was constructed from surveys conducted by the American Community Survey in 2009. The data provides information about a geographical area, including education levels, household income, demographics, household size, gender statistics and age groups. 
The classification task was to predict the geographical location of an area based on sociological and economical parameters of the region. The class label was equal to 1 if the center of the geographical unit had a decimal latitude above 39.09916, which corresponds to the latitude of the $16^{th}$ Circuit Court of Jackson County in Missouri, USA.

\subsection{AMEX}
The dataset was constructed from the daily opening, closing, high and low prices, as well as traded volumes, for stocks at the American Stock Exchange between 1970 and 2010. For each stock, we divided the time series into segments of 10 days.
The task was to identify whether the highest price at the tenth day had a 5\% increase over the highest price at the ninth day, using only information from day 1 to day 9. A class label of one indicated that 
\begin{equation*}
 \frac{\text{high}_{day10} - \text{high}_{day9}}{\text{high}_{day9}} > 1.05.
\end{equation*}

\subsection{ENRON}
This dataset was constructed based on the email exchanges from employees at ENRON. The collection contains emails from about 150 users, mostly senior management of Enron, made public and posted to the web by the Federal Energy Regulatory Commission during its investigation. For each email we constructed a feature space corresponding to the histogram of occurrences of manually selected keywords.
The task was to predict whether an email included email addresses from a domain that is different from enron.com. A class label of 1 indicated that at least one of the addresses in the To, CC or BCC fields of the email contained a different domain than enron.com

\subsection{GEO}
The dataset was constructed from Sea-viewing Wide Field-of-view Sensor (SeaWiFS) data on the Indicators of Coastal Water Quality Collection, originally collected to determine concentrations of chlorophyll-a in the coastal water. The data consists of gridded satellite measurements of chlorophyll-a concentrations (in nanogram/cubic meter) in a band extending between 10 and 100 km from the shoreline \cite{GEO}. The grids are annual composites at a resolution of 5 arc-minutes (approximately 9 x 9 km at the equator). The gridding was done by the Columbia University Center for International Earth Science Information Network (CIESIN). In our dataset, features correspond to measurements from previous years for the same geographical unit, as well as convolution of yearly measurements, the latest being in 2007, using different random kernels of increasing size. 
The task was to predict whether coastal chlorophyll-a increased in 2008 relative to 2007. A class label of 1 indicated that chlorophyll-a indeed increased in 2008 relative to 2007.

\subsection{LASTFM}
The dataset was constructed from tags assigned by listeners to songs broadcast by the online service Last.FM in 2007 (\url{http://musicmachinery.com/2010/11/10/lastfm-artisttags2007/}). We selected the most common 995 tags in the entire dataset and described each song as the histogram of counts for these 995 tags. The task was to identify whether a song was ever tagged, at least once, with a tag containing the word "favorite". A class label of 1 indicated that the song had at least one user assigning a tag containing the word "favorite".

\subsection{NASDAQ}
The dataset was constructed from the daily opening, closing, high and low prices, as well as traded volumes, for stocks at the National Association of Securities Dealers Automated Quotations Stock Exchange between 1970 and 2010. For each stock, we divided the time series into segments of 10 days.
The task was to identify whether the opening price at the tenth day was higher than the closing price at the ninth day, using only information from day 1 to day 9. A class label of one indicated that 
\begin{equation*}
 \text{open}_{day10} > \text{close}_{day9}.
\end{equation*}

\subsection{NYSE}
The dataset was constructed from the daily opening, closing, high and low prices, as well as traded volumes, for stocks at the New York Stock Exchange between 1970 and 2010. For each stock, we divided the time series into segments of 10 days.
The task was to identify whether the highest price at the tenth day had a 5\% increase over the highest price at the ninth day, using only information from day 1 to day 9. A class label of one indicated that 
\begin{equation*}
 \frac{\text{high}_{day10} - \text{high}_{day9}}{\text{high}_{day9}} > 1.05.
\end{equation*}

\subsection{PNS}
The dataset was constructed from a list of common place names. The task was to determine whether the first letter of a place is a vowel, excluding the letter y, based on the histogram of the letters composing the rest of the place name. A class label of 1 indicated that the letter was a vowel. PNS is an acronym for Place Name Strings.

\subsection{SP500}
The dataset was constructed from the daily opening, closing, high and low prices, as well as traded volumes, for S\&P 500 stocks. For each stock, we divided the time series into segments of 8 days.
The task was to identify whether the opening price at the eighth day had an increase over the closing price at the seventh day, using only information from day 1 to day 7. A class label of one indicated that 
\begin{equation*}
 \text{open}_{day8} > \text{close}_{day7}.
\end{equation*}

\newpage
\section{Supplementary Tables}
\begin{table*}[h]
\caption{\textbf{Summary of the datasets.}}\label{SI-tabData}
{\small
\begin{tabular}{l r r c c }
\multicolumn{5}{l}{Datasets from the UCI repository \cite{REF16}} \\\hline 
\bf{Dataset} & \bf{Instances} & \bf{Features} & \bf{Class} & \bf{Reference} \\ \hline
AD (Abalone data) & 4,177 & 8 & male/female & \cite{REF23} \\
ID (Ionosphere data) & 351 & 34 & good return/bad return & \cite{REF21} \\
MGT (MAGIC Gamma Telescope) & 19,020 & 11 & signal/background & \cite{REF20} \\
MM (Mammographic masses) & 961 & 6 & disease severity (2 classes) & \cite{REF22} \\
PD (Parkinson data) & 197 & 23 & affected/unaffected & \cite{REF19} \\
SD (Spambase data) & 4,601 & 57 & spam/not spam & \cite{REF16} \\
WBC (Wisconsin breast cancer data) & 699 & 10 & benign/malignant & \cite{REF18} \\
YBC (Yale breast cancer data) & 650 & 6 & nodal status & \cite{REF17} \\ \hline
\\
\\
\multicolumn{5}{l}{Custom datasets} \\\hline 
\bf{Dataset} & \bf{Instances} & \bf{Features} & \bf{Field} & \bf{Reference} \\ \hline
ACS & 321,583 & 53 & sociology/economy/geography & \cite{ACS} \\
AMEX & 190,769 & 45 & finance & \cite{AMEX} \\
ENRON & 517,424 & 64 & text analysis & \cite{ENRON} \\
GEO & 494,268 & 45 & ecology/geography & \cite{GEO} \\
LASTFM & 20,908 & 995 & recommendation systems & \cite{LASTFM} \\
NASDAQ & 847,427 & 45 & finance & \cite{NASDAQ} \\
NYSE & 919,792 & 45 & finance & \cite{NYSE} \\
PNS & 10,196 & 27 & text analysis & \cite{PNS} \\
SP500 & 15,028 & 35 & finance & \cite{SP500} \\\hline 
\end{tabular}
}
\end{table*}

\begin{table*}[!ht]
\caption{\textbf{Summary of the machine learning classifiers from Weka \cite{REF24}.}}\label{SI-tabMethods}
{\footnotesize
\begin{tabular}{l c l}
\textbf{classifier/meta-learner} & \textbf{Weka class}   & \textbf{Description}  \\ \hline
KNN (k=1, odd) & lazy/IBk & k-nearest neighbor classifier with k=1 \\
KNN (k=2, even) & lazy/IBk & k-nearest neighbor classifier with k=2\\
KNN (k=5) & lazy/IBk & k-nearest neighbor classifier with k=5 \\
k-Star & lazy/KStar & Instance-based learner using entropy-based distance\\
DecisionStump & trees/DecisionStump & One-level decision tree \\
J48 & trees/J48 & Decision tree with pruning\\
REPTree & trees/REPTree & Decision tree using information gain\\
JRip & Rules/JRip &  Propositional rule learner\\
LMT & trees/LMT & Logistic model trees\\
LWL & lazy/LWL & Locally weighted learning algorithm\\
Logistic regression & functions/SimpleLogistic & Logistic regression\\
Regularized Logistic regression & functions/Logistic & Regularized logistic regression\\
Sequential Minimal Optimization & function/SMO & Sequential minimal optimization for SVM\\
NaiveBayes & bayes/NaiveBayes & Na\"ive Bayes classifier\\
M5P & rules/M5P & M5 Model trees and rules\\
OneR & rules/OneR & Minimum-error attribute classifier\\
PART & rules/PART & Partial decision trees classifier \\
RandomForest (n=10 trees) & trees/RandomForest & Random Forest classifier with n=10\\
RandomForest (n=20 trees) & trees/RandomForest & Random Forest classifier with n=20\\
Multilayer Perceptron & functions/MultilayerPerceptron & Multilayer neural network using backpropagation\\
Voted Perceptron & functions/VotedPerceptron & Voted perceptron classifier \\
SGD & functions/SGD & Stochastic gradient descent \\ 
Voting & meta/Vote & Majority voting of an ensemble of J48 classifiers\\ 
Stacking & meta/Stacking & Stacking of an ensemble of J48 classifiers\\ 
AdaBoost + NaiveBayes & meta/AdaBoostM1 & AdaBoost of an ensemble of Na\"ive Bayes  classifiers\\
AdaBoost + Logistic Regression & meta/AdaBoostM1 & AdaBoost of an ensemble of logistic regressions \\
AdaBoost + J48 & meta/AdaBoostM& AdaBoost of an ensemble of J48 classifiers \\
Bagging + REPTree & meta/Bagging & Bagging of an ensemble of REPTrees \\
Bagging + RandomTree & meta/Bagging & Bagging of an ensemble of Random Trees \\
Bagging + RandomForest & meta/Bagging & Bagging of an ensemble of Random Forests \\
LogitBoost + ZeroR & meta/LogitBoost  & ZeroR classifiers use the mode as prediction\\ 
LogitBoost + KNN & meta/LogitBoost & LogitBoost of an ensemble of KNN classifiers\\ 
LogitBoost + DecisionStump & meta/LogitBoost & LogitBoost of an ensemble of Decision Stumps \\ \hline
\end{tabular}
}
\end{table*}

\clearpage

\begin{table*}[!h]
\caption{\textbf{Characteristics of the ensemble of predictions for the real world datasets.} The deviation from the assumption of conditional independence is expressed as the absolute value of the difference $\Delta$ between the two sides of Eq.~\eqref{MT-StatIndependence} in the main text. With the exception of the median true rank of the best inferred predictor, noted as $r_\text{best}$, all quantities are averages over all runs.}\label{SI-TableCondInd}
\begin{tabular}{l c c c c}
\hline
\textbf{Dataset} & \textbf{$|\Delta|$} & $\lambda_1/\sum \lambda$~(\%) & $\lambda_2/\sum \lambda$~(\%) & $r_\text{best}$ \\
\hline
ACS & 0.0067 & 70.0 & 9.0 & 1 \\
AD & 0.0082 & 37.4 & 12.8 & 8 \\
AMEX & 0.0001 & 59.0 & 14.2 & 1 \\
ENRON & 0.0035 & 47.2 & 12.7 & 2 \\
GEO  & 0.0034 & 37.1 & 15.7 & 2\\
ID & 0.0158 & 75.9 & 4.2 & 2 \\
LASTFM  & 0.0063 & 78.1 & 5.4 & 3\\
MGT & 0.0107 & 70.4 & 6.3 & 4 \\
MM & 0.0138 & 80.4 & 5.2 & 3 \\
NASDAQ  & 0.0015 & 50.4 & 13 & 1\\
NYSE  & 0.0001 & 57.3 & 17.3 & 2\\
PD & 0.0101 & 51.4 & 9.8 & 5 \\
PNS & 0.0150 & 57.9 & 23.1 & 2 \\
SD & 0.0059 & 79.8 & 3.5 & 2 \\
SP500 & 0.0016 & 38.8 & 15.7 & 3 \\
WBC & 0.0033 & 92.8 & 1.4 & 2 \\
YBC & 0.0209 & 50.0 & 10.2 & 4 \\
\end{tabular}
\end{table*}

\newpage
\section{Supplementary Figures}

\begin{figure}[!ht]
\centering
\epsfig{file=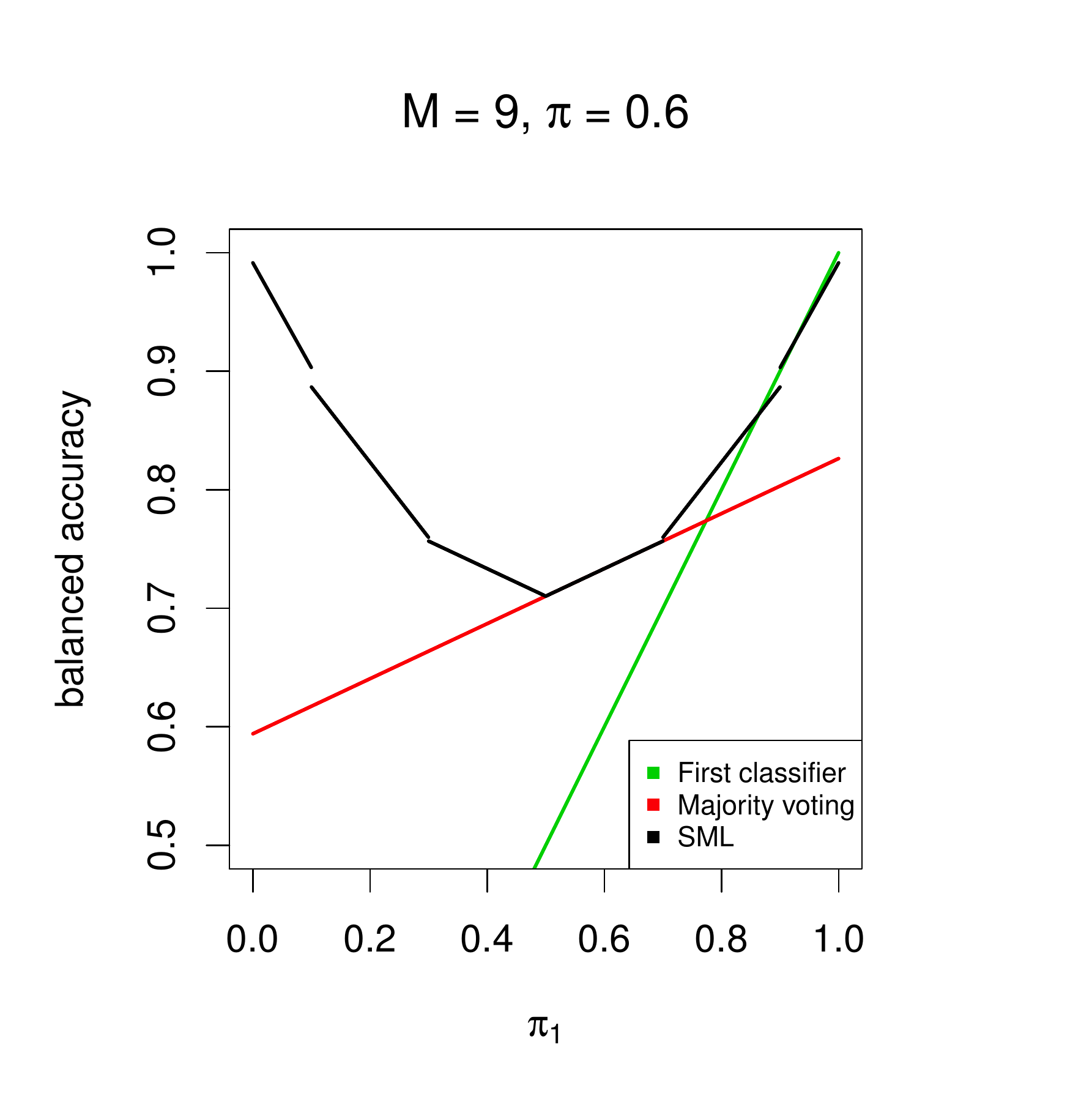,width=\linewidth}
\caption{Performance of the first algorithm, majority voting and SML as a function of the balanced accuracy of the first algorithm when all other algorithms in the ensemble have identical sensitivities $\pi = 0.6$. In this illustrative example, $M=9$. The performance of majority voting (in red) changes linearly with $\pi_1$, albeit with partial derivative smaller than 1. The balanced accuracy of SML (in black) is a piecewise linear function of $\pi_1$. The jumps in the balanced accuracy of SML occur when the value $(M-1)/2 - 1/2\cdot 1/(2\pi - 1)$ is an integer.}
\label{SI-fig:lemmaExample}
\end{figure}

\begin{figure}[!ht]
\centering
\epsfig{file=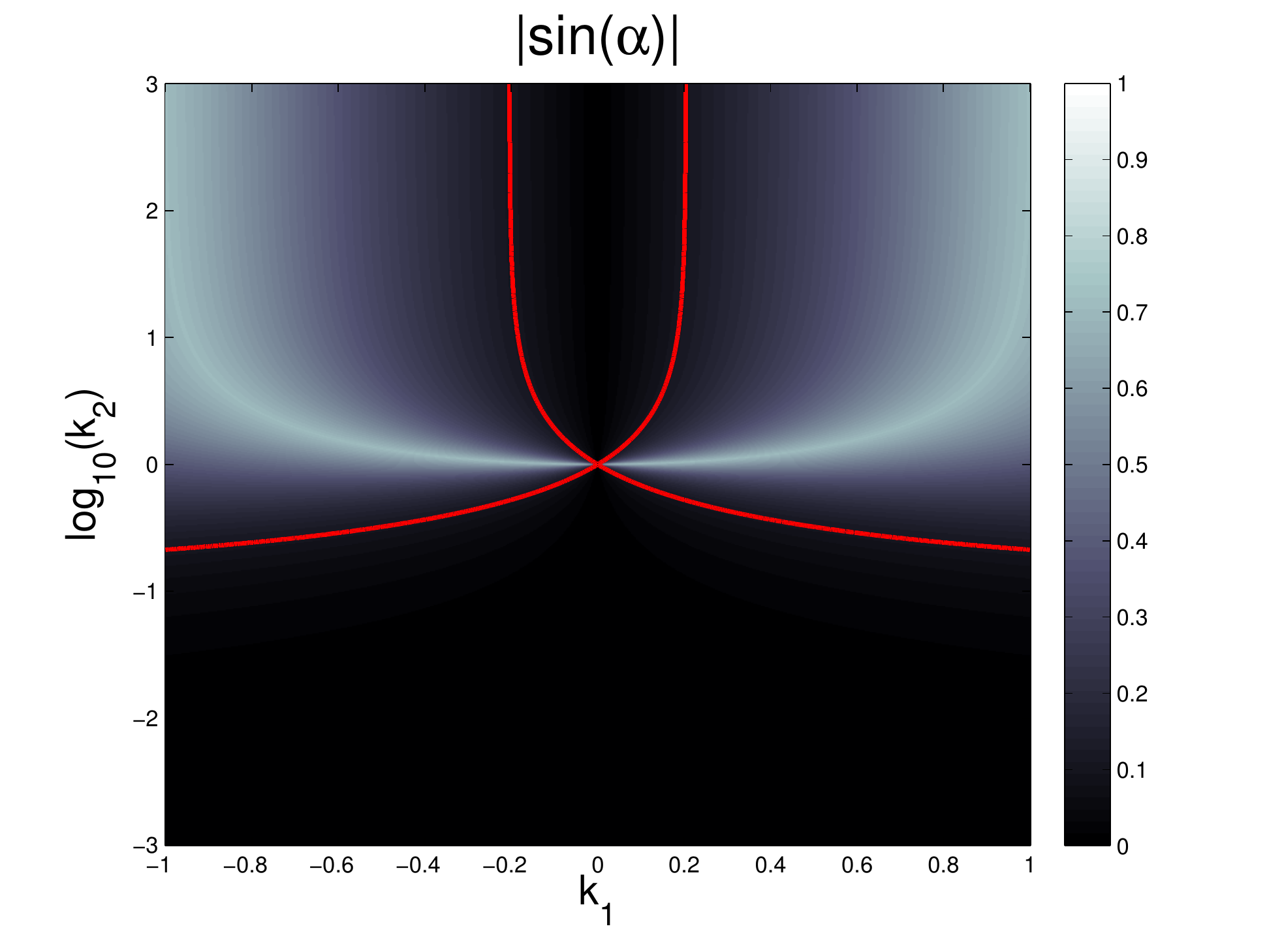,width=\linewidth}
\caption{The heatmap shows the absolute value of the angle between the truth and the eigenvector $e_1$, on which the SML prediction is based. The dark area between the two red lines graphically shows the relationship between $k_1$ and $k_2$ such that $| \alpha | \leq 6^\circ$. The figure shows that SML is robust to cartels: when $\alpha \approx 0$, the honest classifiers lie approximatively on the eigenvector $e_1$.}
\label{SI-fig_FIG1}
\end{figure}

\clearpage

\begin{figure}[!ht]
\centering
\epsfig{file=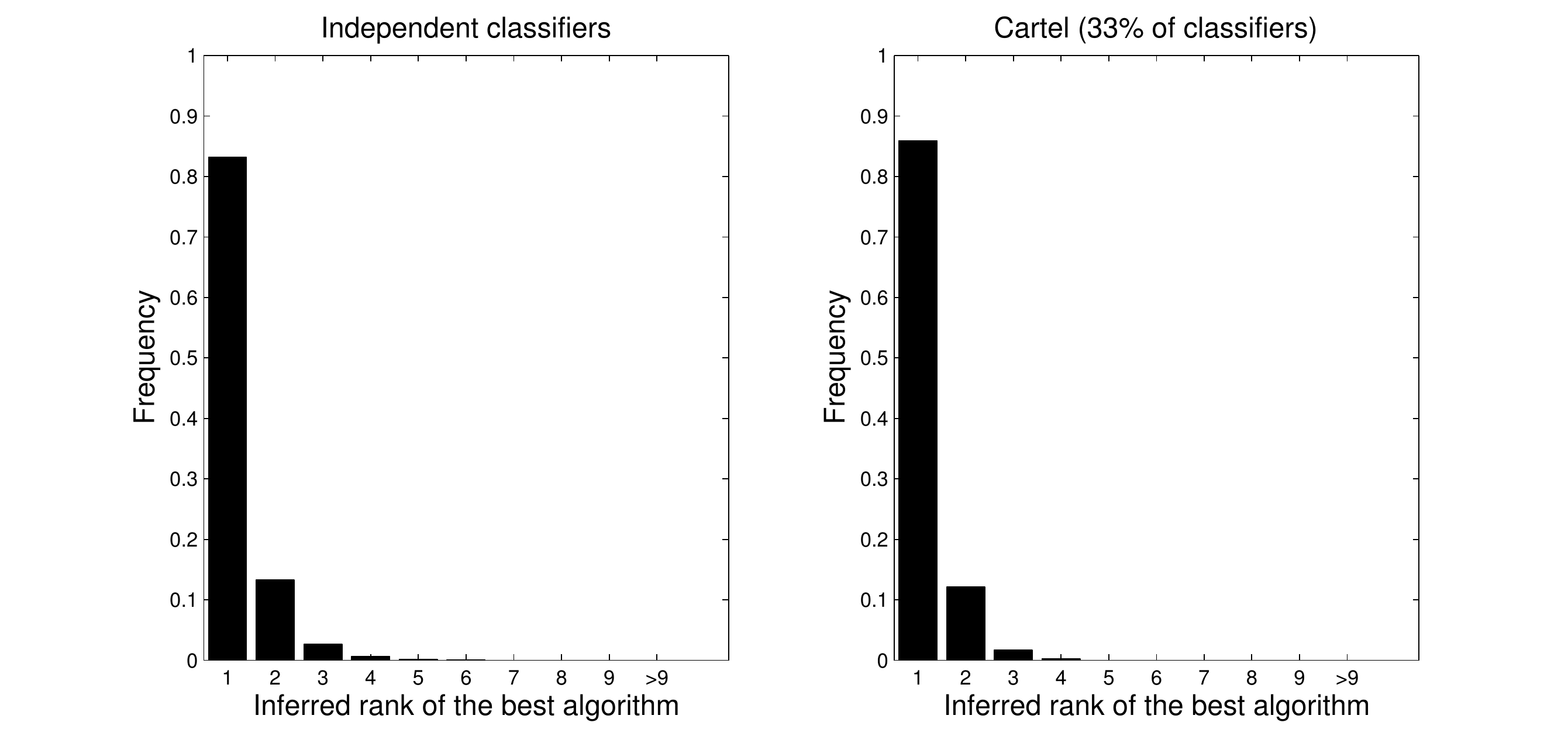,width=\linewidth}
\caption{The largest entry in the leading eigenvector often corresponds to the best classifier in the ensemble. In the plots, each bar represents the empirical probability that the entry in the leading eigenvector corresponding to best classifier attained a specific rank.}
\label{SI-fig_FIG2}
\end{figure}

\vfill

\begin{figure}[!ht]
\centering
\epsfig{file=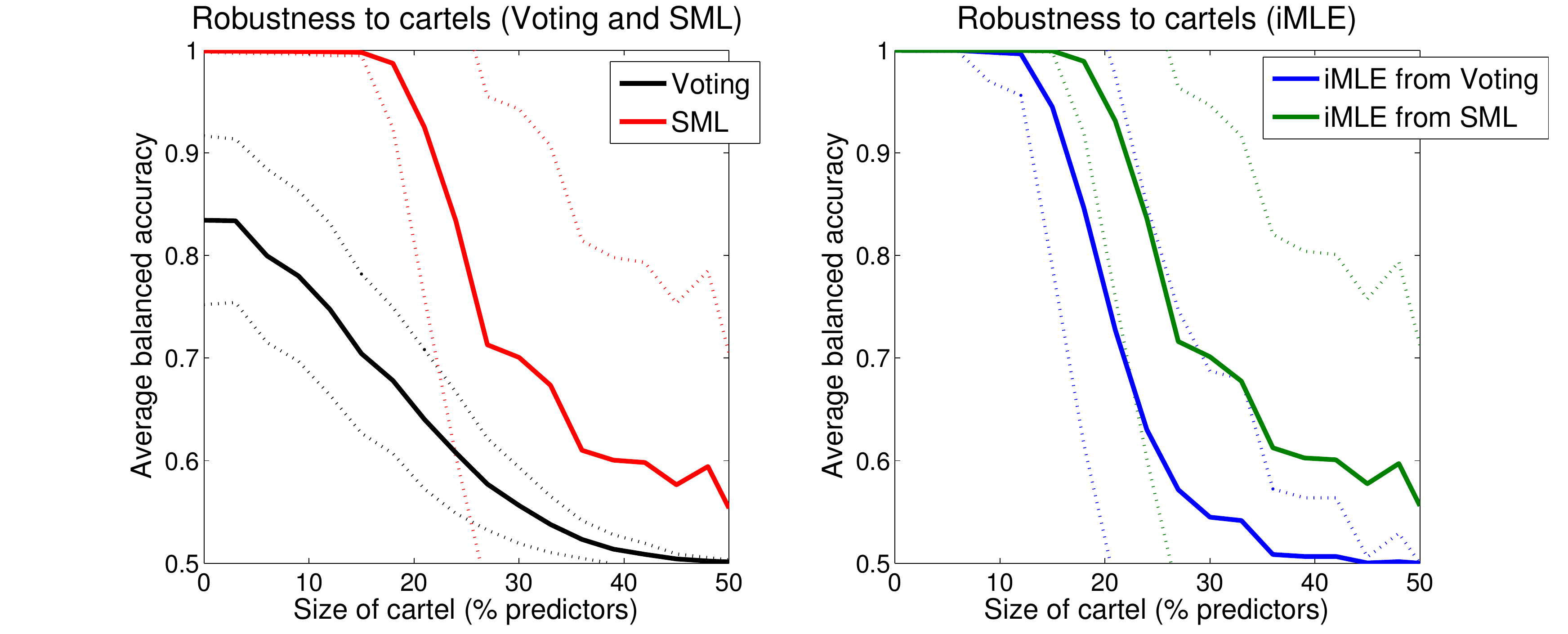,width=\linewidth}
\caption{SML is more robust to cartels than majority voting (left panel). iMLE using SML estimates as starting point is also more robust to cartels than iMLE using majority voting as the starting condition (right panel). For each meta-learner prediction the average balanced accuracy is shown (filled lines) together with the standard error (dotted lines, n=500 runs for each cartel's fraction).}\label{SI-fig_FIG6}
\end{figure}

\begin{figure}[!ht]
\centering
\epsfig{file=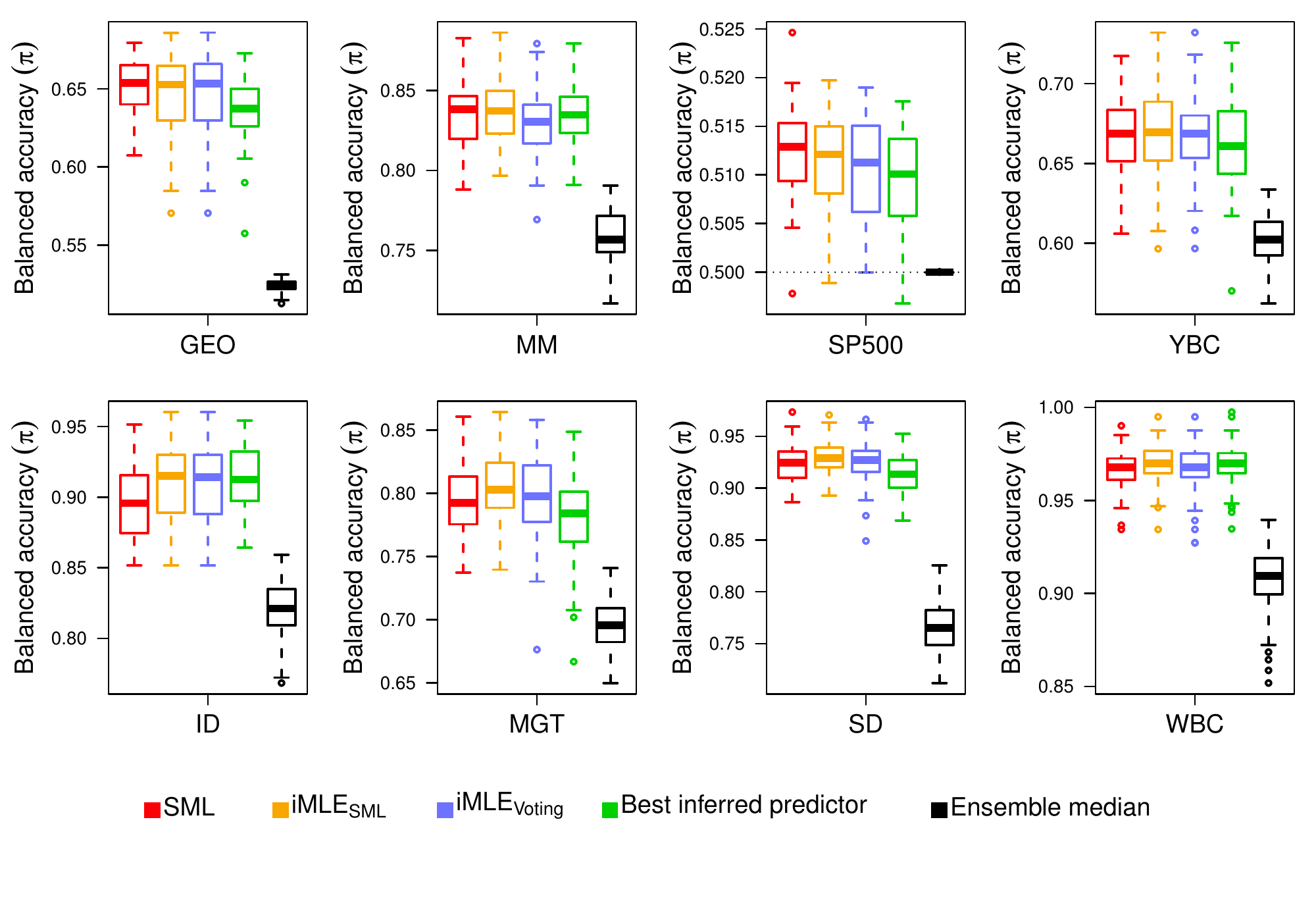,width=\linewidth}
\caption{Comparison of  several classifiers on real-world datasets where our conditions are nearly satisfied. The median balanced accuracy of all classifiers in the ensemble is shown in black.}
\label{SI-fig_FIG10}
\end{figure}

\begin{figure}[!ht]
\centering
\epsfig{file=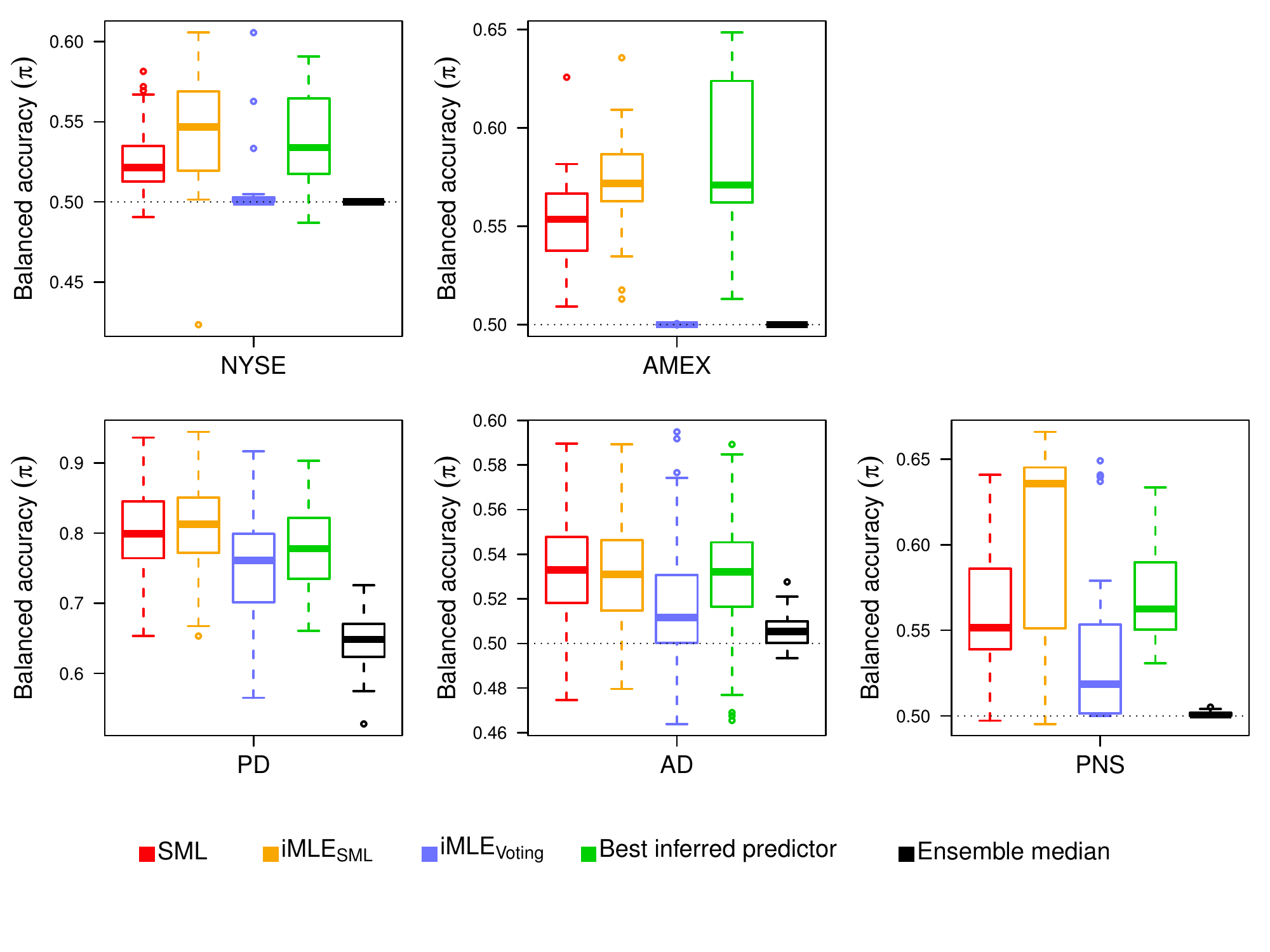,width=\linewidth}
\caption{Comparison of  several classifiers on real-world datasets, where predictors have structure similar to that of cartels. The median balanced accuracy of all classifiers in the ensemble is shown in black.}
\label{SI-fig_FIG11}
\end{figure}

\begin{figure}[!ht]
\centering
\epsfig{file=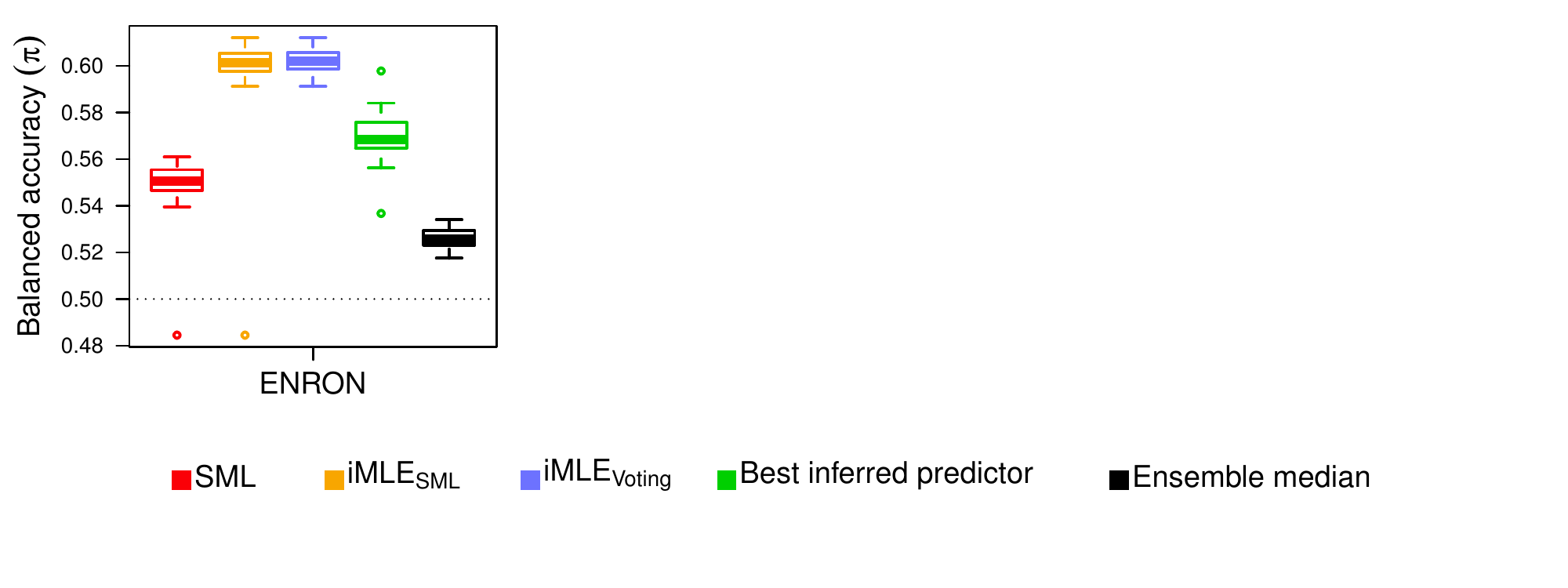,width=\linewidth}
\caption{Comparison of  several classifiers on the ENRON dataset, characterized by a sparse feature space. The median balanced accuracy of all classifiers in the ensemble is shown in black.}
\label{SI-fig_FIG12}
\end{figure}

\newpage
\clearpage

\section{Supplementary References}
\begingroup
\renewcommand{\addcontentsline}[3]{}
\renewcommand{\section}[2]{}

\endgroup


\begin{thebibliography}{10}

\bibitem{tsai_combining_2010}
Tsai, C.-F \& Hsiao, Y.-C.
\newblock (2010) Combining multiple feature selection methods for stock prediction: Union, intersection, and multi-intersection approaches {\em Decis. Support Syst.} {\bf 50}, 258-269.

\bibitem{zweig_making_2011}
Zweig, J.
\newblock (2011) Making sense of market forecasts, {\em Wall Street Journal - Jan 8, 2011}.

\bibitem{european_central_bank_survey}
{European Central Bank}.
\newblock Survey of professional forecasters.
\newline\url{www.ecb.int/stats/prices/indic/forecast/html/index.en.html}

\bibitem{federal_reserve_bank_of_philadelphia_survey}
{Federal Reserve Bank of Philadelphia}.
\newblock Survey of professional forecasters.
www.philadelphiafed.org/research-and-data/real-time-center/survey-of-professional-forecasters/

\bibitem{monetary_authority_of_singapore_survey}
{Monetary Authority of Singapore}.
\newblock Survey of professional forecasters.
www.mas.gov.sg/news-and-publications/surveys/mas-survey-of-professional-forecasters.aspx

\bibitem{genre_combining_2010}
Genre, V, Kenny, G, Meyler, A,  \& Timmermann, A.~G.
\newblock (2010) Combining the forecasts in the {ECB} survey of professional
  forecasters: Can anything beat the simple average?, (Social Science Research
  Network, Rochester, {NY}), {SSRN} Scholarly Paper {ID} 1719622.

\bibitem{micsinai_picking_2012}
Micsinai, M, et al.
\newblock (2012) Picking ChIP-seq peak detectors for analyzing chromatin modification experiments {\em Nucleic acids research} {\bf 40}, e70.
\newblock {PMID:} 22307239.

\bibitem{wright_multidisciplinary_2007}
Wright, F, De~Vito, C, Langer, B,  \& Hunter, A.
\newblock (2007) Multidisciplinary cancer conferences: A systematic review and development of practice standards {\em European Journal of Cancer} {\bf 43}, 1002--1010.

\bibitem{cole_peer_1978}
Cole, S, et al.
\newblock (1978) Peer review in the national science foundation: Phase one of a
  study., (Office of Publications, National Academy of Sciences, 2101
  Constitution Ave., {N.W.}, Washington, {D.C.} 20418 ({ISBN}
  0-309-02788-8))

\bibitem{margolin_systematic_2013}
Margolin, A.~A, et al.
\newblock (2013) Systematic Analysis of Challenge-Driven Improvements in Molecular Prognostic Models for Breast Cancer {\em Science Translational Medicine} {\bf 5},~181

\bibitem{linstone_delphi}
Linstone, H \& Turoff, M.~e.
\newblock (1975) {\em Delphi Method : Techniques and Applications}.
\newblock (Addison-Wesley Pub. Co)

\bibitem{stock_forecasting_2006}
Stock, J.~H \& Watson, M.~W.
\newblock (2006) Forecasting with many predictors, (Elsevier), Handbook of
  economic forecasting.

\bibitem{timmermann_chapter_2006}
Timmermann, A.
\newblock (2006) in {\em Handbook of Economic Forecasting}, eds.{} G.~Elliott,
  C.~G \& Timmermann, A.
\newblock (Elsevier) Volume 1, pp. 135--196.

\bibitem{DeGroot}
Degroot, MH, 
\newblock{(1974) Reaching a concensus, J. Amer. Stat. Assoc., 69, 118-121.}

\bibitem{dawid_maximum_1979}
Dawid, A.~P \& Skene, A.~M.
\newblock (1979) {\em Journal of the Royal Statistical Society. Series C
  (Applied Statistics)} {\bf 28}, 20--28.

\bibitem{karger_budget-optimal_2011}
Karger, D.~R, Oh, S,  \& Shah, D.
\newblock (2011) {\em Budget-optimal crowdsourcing using low-rank matrix
  approximations}.
\newblock {\em Proc. of the IEEE Allerton Conf. on Communication, Control, and Computing}, pp. 284--291.

\bibitem{karger_iterative_2011}
Karger, D.~R, Oh, S,  \& Shah, D.
\newblock (2011) Iterative Learning for Reliable Crowdsourcing Systems
 {\em {NIPS} 1953-1961}.

\bibitem{whitehill_whose}
Whitehill, J, Ruvolo, P, Wu, T, Bergsma, J,  \& Movellan, J.
\newblock (2009) {\em Whose Vote Should Count More: Optimal Integration of
  Labels from Labelers of Unknown Expertise}.
\newblock {\em {NIPS}}.

\bibitem{padhraic_smyth_inferring_1996}
Padhraic~Smyth, U.~F.
\newblock (1996) Inferring Ground Truth from Subjective Labelling of Venus Images

\bibitem{sheng_get_2008}
Sheng, V.~S, Provost, F,  \& Ipeirotis, P.~G.
\newblock (2008) {\em Get another label? improving data quality and data mining
  using multiple, noisy labelers}, {KDD} '08.
\newblock ({ACM}, New York, {NY}, {USA}), p. 614–622.

\bibitem{welinder_multidimensional_2010}
Welinder, P, Branson, S, Belongie, S,  \& Perona, P.
\newblock (2010) in {\em Advances in Neural Information Processing Systems 23},
  eds.{} Lafferty, J, Williams, C, Shawe-Taylor, J, Zemel, R,  \& Culotta, A.
\newblock pp. 2424--2432.

\bibitem{yan_modeling_2010}
Yan, Y, et al.
\newblock (2010) {\em Modeling annotator expertise: Learning when everybody
  knows a bit of something}.
\newblock (AISTATS 2012) pages 932–939, 2010

\bibitem{raykar_learning_2010}
Raykar, V.~C, et al.
\newblock (2010) Learning From Crowds
{\em J. Mach. Learn. Res.} {\bf 11}, 1297–1322.

\bibitem{dietterich_ensemble_2000}
Dietterich, T.~G.
\newblock (2000) {\em Ensemble Methods in Machine Learning}.
\newblock (Springer), p. 1–15.

\bibitem{witten_data_2011}
Witten, I.~H, Frank, E,  \& Hall, M.~A.
\newblock (2011) {\em Data mining: practical machine learning tools and
  techniques}.
\newblock (Morgan Kaufmann, Burlington, {MA}).

\bibitem{jin_learning_2003}
Jin, R \& Ghahramani, Z.
\newblock (2003) {\em Learning with Multiple Labels}.
in: S. Becker, S. Thrun, K. Obermayer (Eds.), Advances in Neural Information Processing Systems, vol. 15, MIT Press, Cambridge, MA, 2003, pp. 897–904.

\bibitem{lauritzen_em_1995}
Lauritzen, S.~L.
\newblock (1995) "The EM algorithm for graphical association models with missing data." Computational Statistics \& Data Analysis 19.2: 191-201.

\bibitem{snow_cheap_2008}
Snow, R, {O'Connor}, B, Jurafsky, D,  \& Ng, A.~Y.
\newblock (2008) {\em Cheap and fast—but is it good?: evaluating non-expert
  annotations for natural language tasks}, {EMNLP} '08.
\newblock (Association for Computational Linguistics, Stroudsburg, {PA},
  {USA}), p. 254–263.

\bibitem{walter_estimation_1988}
Walter, S.~D \& Irwig, L.~M.
\newblock (1988) Estimation of test error rates, disease prevalence and relative risk from misclassified data: a review
{\em Journal of clinical epidemiology} {\bf 41}, 923--937.
\newblock {PMID:} 3054000.


\bibitem{strino_vda_2011}
Strino, F, Parisi, F,  \& Kluger, Y.
\newblock (2011) VDA, a method of choosing a better algorithm with fewer validations
{\em {PloS} one} {\bf 6}, e26074.
\newblock {PMID:} 22046256.

\end{thebibliography}

\begin{thebibliography}{10}{}
\bibitem{candes_exact_2009}
Candes, E.~J \& Recht, B.
\newblock (2009) 
{Exact matrix completion via convex optimization,} 
{\em Foundations of Computational Mathematics} {\bf 9},
  717--772.

\bibitem{karger_budget-optimal_2011_SI}
Karger, D.~R, Oh, S,  \& Shah, D.
\newblock (2011) 
{Budget-optimal crowdsourcing using low-rank matrix approximations}.
{\em Proc. of the IEEE Allerton Conf. on Communication, Control, and Computing},
pp. 284--291.

\bibitem{Kato}
Kato, T. (1995). 
{\em Perturbation theory for linear operators}, 
2nd edition, Springer-Verlag. 

\bibitem{REFVDA}
F.~Strino, F.~Parisi, and Y.~Kluger.
\newblock VDA, a method of choosing a better classifier with fewer validations.
\newblock {\em PLoS ONE}, 6(10):e26074, 2011.

\bibitem{REF23}
P.~McShane and J.~Reyn.
\newblock Small-scale spatial variation in growth, size at maturity, and
  yield-and egg-per-recruit relations in the new zealand Abalone Haliotis 
\newblock {\em New Zealand Journal of Marine and Freshwater Research},
  29(4):603--612, 1995.

\bibitem{REF21}
V.~Sigillito, S.~Wing, L.~Hutton, and K.~Baker.
\newblock Classification of radar returns from the ionosphere using neural
  networks.
\newblock {\em Johns Hopkins APL Technical Digest}, 10(3):262--266, 1989.

\bibitem{REF20}
D.~Heck, J.~Knapp, J.~Capdevielle, G.~Schatz, and T.~Thouw.
\newblock Report FZKA 6019, Forschungszentrum Karlsruhe, 1998.
\newblock Technical report, 1986.

\bibitem{REF22}
M.~Elter, R.~Schulz-Wendtland, and T.~Wittenberg.
\newblock The classification of breast cancer biopsy outcomes using two 
  approaches that both emphasize an intelligible decision process.
\newblock {\em Med Phys}, 34(11):4164--72, Nov 2007.

\bibitem{REF19}
M.~A. Little, P.~E. McSharry, S.~J. Roberts, D.~A.~E. Costello, and I.~M.
  Moroz.
\newblock Exploiting nonlinear recurrence and fractal scaling properties for voice disorder detection.
\newblock {\em Biomed Eng Online}, 6:23, 2007.

\bibitem{REF16}
A. Asuncion and D.J. Newman.
\newblock {UCI} machine learning repository. Irvine, CA: University of California, Department of Information and Computer Science. 
\url{http://www.ics.uci.edu/~mlearn/{MLR}epository.htm}

\bibitem{REF18}
W.~H. Wolberg and O.~L. Mangasarian.
\newblock Multisurface method of pattern separation for medical diagnosis
  applied to breast cytology.
\newblock {\em Proc Natl Acad Sci U S A}, 87(23):9193--6, Dec 1990.

\bibitem{REF17}
F.~Parisi, A.~M. Gonz{\'a}lez, Y.~Nadler, R.~L. Camp, D.~L. Rimm, H.~M. Kluger,
  and Y.~Kluger.
\newblock Benefits of biomarker selection and clinico-pathological covariate
  inclusion in breast cancer prognostic models.
\newblock {\em Breast Cancer Res}, 12(5):R66, Sep 2010.


\bibitem{ACS}
Data prepared by Infochimps 
\newblock {\em US Census (ACS): Income, Age, Housing and Population by Location (2009) }
\newline\url{http://www.infochimps.com/datasets/us-census-acs-income-age-housing-and-population-by-location}

\bibitem{AMEX}
Data prepared by Infochimps 
\newblock {\em AMEX Daily 1970-2010 Open, Close, High, Low and Volume }
\newline\url{http://www.infochimps.com/datasets/amex-exchange-daily-1970-2010-open-close-high-low-and-volume}

\bibitem{ENRON}
William W. Cohen
\newblock {\em Enron Email Dataset }
\newblock\url{http://www.cs.cmu.edu/~enron/}

\bibitem{GEO}
Goddard Space Flight Center (GSFC), and Center for International Earth Science Information Network (CIESIN)/Columbia University. 
\newblock {\em Indicators of Coastal Water Quality: Annual Chlorophyll-a Concentration 1998-2007}
\newblock Palisades, NY: NASA Socioeconomic Data and Applications Center (SEDAC). \newline\url{http://sedac.ciesin.columbia.edu/data/set/icwq-annual-chlorophyll-a-concentration-1998-2007}

\bibitem{LASTFM}
Paul Lamere
\newblock {The LastFM-ArtistTags2007 Data set}
\newline\url{http://static.echonest.com/Lastfm-ArtistTags2007.tar.gz}

\bibitem{NASDAQ}
Data prepared by Infochimps 
\newblock {\em NASDAQ Exchange Daily 1970-2010 Open, Close, High, Low and Volume }
\newline\url{http://www.infochimps.com/datasets/nasdaq-exchange-daily-1970-2010-open-close-high-low-and-volume}

\bibitem{NYSE}
Data prepared by Infochimps 
\newblock {\em NYSE Daily 1970-2010 Open, Close, High, Low and Volume }
\newline\url{http://www.infochimps.com/datasets/nyse-daily-1970-2010-open-close-high-low-and-volume}

\bibitem{PNS}
Data prepared by Infochimps 
\newblock {\em Word List - 10,000+ Common Place Names }
\newline\url{http://www.infochimps.com/datasets/word-list-10000-common-place-names}

\bibitem{SP500}
Data prepared by StockWiz, 
\newblock {\em Historical Data for S\&P 500 Stocks }
\newline\url{http://pages.swcp.com/stocks/}

\bibitem{REF24}
I.~Witten, E.~Frank, and M.~Hall.
\newblock {\em Data Mining: Practical machine learning tools and techniques}.
\newblock Morgan Kaufmann, 2011.

\end{thebibliography}
\end{document}